\documentclass[times, review, 10pt]{elsarticle}
\biboptions{sort&compress}

\usepackage{amssymb}
\usepackage{amsmath}
\usepackage{amsthm}
\usepackage{xcolor}
\usepackage{algorithm}
\usepackage{algpseudocode}
\usepackage{hyperref}
\usepackage{multirow}
\usepackage{booktabs}
\usepackage{enumitem}

\newcommand\norm[1]{\left\lVert#1\right\rVert}
\newcommand{\argmin}{\operatornamewithlimits{argmin}}

\usepackage{lineno}

\begin{document}

\begin{frontmatter}



\title{Variable Selection Using Relative Importance Rankings}


\author{Tien-En Chang} 
\author{Argon Chen}

\affiliation{organization={Institute of Industrial Engineering, National Taiwan University},
            addressline={No.1, Sec. 4, Roosevelt Road}, 
            city={Taipei},
            postcode={10617}, 
            country={Taiwan}}

\begin{abstract}
Although conceptually related, variable selection and relative importance (RI) analysis have been treated quite differently in the literature. While RI is typically used for post-hoc model explanation, this paper explores its potential for variable or feature ranking and filter-based selection before model creation. Specifically, we anticipate strong performance from the RI measures because they incorporate both direct and combined effects of predictors, addressing a key limitation of marginal correlation, which ignores dependencies among predictors. We implement and evaluate the RI-based variable ranking and selection methods, including a newly proposed RI measure, CRI.Z, with improved computational efficiency relative to conventional RI measures.

Through extensive simulations, we first demonstrate how the RI measures more accurately rank the variables than the marginal correlation, especially when there are suppressed or weak predictors. We then show that predictive models built on these rankings are highly competitive, often outperforming state-of-the-art linear-model methods such as the lasso and relaxed lasso. The proposed RI-based methods are particularly effective in challenging cases involving clusters of highly correlated predictors, a setting known to cause failures in many benchmark methods. The practical utility and efficiency of RI-based methods are further demonstrated through two high-dimensional gene expression datasets. Although lasso methods have dominated the recent literature on variable selection, our study reveals that the RI-based method is a powerful and competitive alternative. We believe these underutilized tools deserve greater attention in statistics and machine learning communities. The code is available at: \url{https://github.com/tien-endotchang/RI-variable-selection}.
\end{abstract}



\begin{keyword}
Relative importance analysis \sep variable selection \sep variable ranking

\end{keyword}

\end{frontmatter}



\section{Introduction}\label{sec:intro}
Variable selection, also referred to as feature selection, is a fundamental problem in statistics and machine learning. Its primary goal is to identify a subset of variables with substantive predictive relevance from a larger candidate set, enabling the construction of parsimonious, interpretable, and robust models \citep{guyon2003introduction}. This task is particularly challenging in high-dimensional settings, where the number of predictors $p$ far exceeds the number of observations $n$. A well-known example is identifying cancer-related genes from microarray data, where thousands of gene expressions are measured for fewer than a hundred patients \citep{golub1999molecular,deng2013gene,meenachi2021metaheuristic,shen2020comprehensive}.

Many approaches have been introduced to address this challenge. They are typically categorized into wrappers, embedded, and filter methods \cite{guyon2003introduction}. Wrappers, such as best subset selection \cite{beale1967discarding,hocking1967selection} and its greedy alternative, forward stepwise selection \cite{efroymson1966stepwise,draper1998applied}, use model performance to evaluate candidate subsets. In recent years, a variety of meta-heuristic optimization algorithms have also been developed for wrapper methods \citep{de2020binary,meenachi2021metaheuristic,gao2025quantum}. Embedded methods, such as the lasso \citep{tibshirani1996regression}, incorporate variable selection directly into model training and have become dominant in the literature, attracting tens of thousands of citations. Extensions of tree-based approaches have also been proposed as embedded methods \citep{deng2013gene}. Filter methods, in contrast, decouple variable ranking from model fitting. While various ranking strategies have been proposed, such as those based on entropy \citep{zou2026feature}, the most widely used approach is Sure Independence Screening (SIS) \citep{fan2008sure}, which ranks predictors by their marginal correlation with the response. Although computationally efficient, SIS is limited by its reliance on marginal correlations, which can be misleading when predictors are correlated, a common feature of real-world data. In such scenarios, the predictive power overlaps among predictors, making it difficult to select relevant features for best prediction \cite{fan2008sure}. For instance, in genomics, where gene expression levels frequently exhibit severe multicollinearity, SIS has been shown to overlook critical genes for leukemia classification \cite{hong2016data}. 

A related yet conceptually distinct problem is to assess predictor importance in the presence of multicollinearity. Originating in quantitative behavioral and psychological research, relative importance (RI) analysis seeks to quantify each variable's unique contribution to the explanatory power of a model \cite{johnson2004history,gromping2015variable}. Unlike marginal correlation or regression coefficient, RI measures such as General Dominance (GD) \citep{budescu1993dominance,azen2003dominance} and Relative Weight (RW) \citep{genizi1993decomposition,johnson2000heuristic} consider both direct effect and combined effects of predictors in the linear model, thus handling the dependencies among predictors \citep{johnson2004history}. Historically, these methods were developed as post-hoc explanatory tools, and some studies have cautioned against their use for variable selection \cite{budescu1993dominance,johnson2000heuristic,tonidandel2011relative,johnson2017best}. Although recent methodological developments \citep{zuber2011high, shen2020comprehensive} and applied studies \citep{zeng2025interpretableA,zeng2025interpretableB} have begun to challenge this position, a systematic evaluation of RI-based variable selection methods remains lacking.

This paper aims to bridge this gap between RI analysis and variable selection. With the estimation of each variable's unique contribution to model explanation, RI measures are expected to offer a robust foundation for filter-based variable selection. In this paper, we evaluate the performance of the established RI measures (GD, RW, CRI \citep{shen2020comprehensive}) in variable selection and model prediction. In addition, we propose a computation-efficient RI-based selection method referred to as CRI.Z. Through extensive simulations and real-world dataset examples, we demonstrate that RI-based selection is not only competitive with modern linear-model benchmarks such as the lasso and relaxed lasso and non-convex penalties, but often superior in scenarios involving high predictor correlation. Our main contributions are:
\begin{itemize}[noitemsep, topsep=0pt]
\item We formalize  a class of filter methods based on RI rankings, and systematically evaluate their performance relative to each other and to simpler methods such as marginal correlation (Section \ref{sec:relative importance} and \ref{sec:simulation part I}). 
\item We propose CRI.Z, a novel and relatively efficient RI ranking method derived via the framework of CRI (Section \ref{subsec: CAR and CRI.Z}).
\item We use the extensive simulations from the variable selection literature to compare the RI-based methods with leading variable selection benchmarks. We demonstrate that RI-based methods are not only competitive but can also outperform modern linear-model benchmarks under specific conditions. (Section \ref{sec:simulation part II}). Two examples of gene expression datasets are also used to demonstrate the utility of the proposed methods. (Section \ref{sec:real-world})
\end{itemize}
\section{Benchmark Variable Selection Methods}\label{sec:variable selection}
This section reviews variable selection methods based on linear models that serve as primary benchmarks for our proposed method. We begin with two classic wrapper methods: best subset and forward stepwise selection. We then review the lasso, the most prominent embedded method, and its variant relaxed lasso. Finally, we describe Sure Independence Screening (SIS), a simple yet widely used filter method. 

We consider the standard linear model, where the response vector $y\in\mathbb{R}^n$ is modeled using a predictor matrix $X\in\mathbb{R}^{n\times p}$, true coefficients $\beta_0\in\mathbb{R}^p$ and noise $\epsilon\in\mathbb{R}^n$ that are independent $N(0,\sigma^2)$:
\begin{equation}
    \label{eq:linear model}
    y=X\beta_0+\epsilon.
\end{equation}
Let $\Sigma\in\mathbb{R}^{p\times p}$ denote the covariance matrix of predictors. The Signal-to-Noise Ratio (SNR) is defined as $\text{SNR}=\beta_0^\top\Sigma\beta_0/\sigma^2$. Throughout this paper, we assume that both the response $y$ and each predictor $x_i$ are standardized to have zero mean and unit $\ell_2$-norm.
\subsection{Best Subset Selection}
Best subset selection \cite{beale1967discarding,hocking1967selection} seeks the model with the best in-sample fit for a given model size $k$. It identifies a subset consisting of $k$ predictors that minimizes the residual sum of squares. This can be formulated as the following non-convex optimization problem:
\begin{equation}
    \label{eq:bs_prob}
    \min_{\beta\in\mathbb{R}^p} \norm{y-X\beta}^2_2\quad\text{subject to}\quad\norm{\beta}_0\leq k,
\end{equation}
where $\norm{\beta}_0=\sum_{i=1}^p\mathbf{1}\{\beta_i\neq 0\}$ denotes the $\ell_0$ norm of $\beta$. While best subset often performs well in high-SNR settings by accurately recovering true signals, it tends to overfit as the SNR is low by selecting spurious predictors \cite{hastie2020best}. Moreover, its primary drawback is computational complexity. The underlying problem is NP-hard \cite{natarajan1995sparse}. Although modern mixed-integer optimization (MIO) solvers have made best subset more practical for moderate-sized datasets \citep{bertsimas2016best}, it remains computationally demanding at scale \cite{hastie2020best}.
\subsection{Forward Stepwise Selection}
Forward stepwise selection \cite{efroymson1966stepwise,draper1998applied} is a greedy approximation to best subset. It builds a model iteratively by adding the predictor that offers the greatest reduction in residual sum of squares. The procedure starts from an empty active set $A_0=\{\}$. At each step $k=1,\ldots,\min\{n,p\}$, the algorithm selects the predictor indexed by $j_k$ as follows:
\begin{equation}
\label{eq:fs}
j_k = \argmin_{j \notin A_{k-1}} \, \norm{y - P_{A_{k-1} \cup \{j_k\}} y}_2^2,
\end{equation}
where $A_{k-1}$ denotes the active set from the previous step and $P_\mathcal{S} y$ denotes the projection of $y$ onto the column space of the predictors indexed by the subset $\mathcal{S}$. The active set is then updated via $A_k=A_{k-1}\cup\{j_k\}$. Forward stepwise typically performs similarly to best subset \citep{hastie2020best}. However, it is far more computationally tractable.

\subsection{The Lasso and Relaxed Lasso}\label{subsec:lasso and relatives}
The lasso \cite{tibshirani1996regression} is one of the most influential methods for variable selection in high-dimensional regression. It provides a convex relaxation of Eq.~\eqref{eq:bs_prob} by replacing the non-convex $\ell_0$ norm with the convex $\ell_1$ norm:
\begin{equation}
    \label{eq:lasso_prob_constrained}
    \min_{\beta\in\mathbb{R}^p} \norm{y-X\beta}^2_2\quad\text{subject to}\quad\norm{\beta}_1\leq t,
\end{equation}
where $t\geq 0$ is a tuning parameter that constrains the $\ell_1$ norm of the estimated coefficients. Equivalently, the penalized form of the lasso is
\begin{equation}
    \label{eq:lasso_prob_penalized}
    \min_{\beta\in\mathbb{R}^p} \norm{y-X\beta}^2_2+\lambda\norm{\beta}_1,
\end{equation}
where $\lambda\geq 0$ is a tuning parameter that controls the regularization strength. The $\ell_1$ penalty induces sparsity by shrinking some coefficients exactly to zero, enabling simultaneous variable selection and coefficient shrinkage. This shrinkage introduces bias but can reduce variance, especially beneficial in low SNR settings where best subset and forward stepwise tend to overfit \citep{hastie2020best}. However, in high SNR settings, this shrinkage can excessively weaken large coefficients, reducing model accuracy.

The relaxed lasso \citep{hastie2020best} addresses this limitation through a two-stage procedure. First, the lasso identifies an active set $A_\lambda$ for a given $\lambda$. Then, a new solution that is computed as a convex combination of the lasso fit and a least squares fit on the active predictors. The estimator is
\begin{equation}
    \label{eq:relaxed lasso}
    \hat{\beta}^{\mathrm{relax}}(\lambda,\gamma)=\gamma\hat\beta^{\mathrm{lasso}}(\lambda)+(1-\gamma)\hat{\beta}^{\mathrm{LS}}_{A_\lambda},
\end{equation}
where $\hat{\beta}^{\mathrm{lasso}}(\lambda)$ is the lasso solution and $\hat{\beta}^{\mathrm{LS}}_{A_\lambda}$ is the least squares fit on the set of variables selected with the penalty $\lambda$ denoted $A_\lambda$. The second tuning parameter, $\gamma\in[0,1]$, allows the model to retain selected variables from the lasso while reducing coefficient shrinkage. The relaxed lasso has demonstrated strong empirical performance and is considered a crucial benchmark in modern variable selection \citep{hastie2020best}. 

Other advanced approaches to mitigating the bias inherent in lasso coefficients involve replacing the $\ell_1$ penalty with non-convex penalties, such as SCAD \citep{fan2001variable} and MCP \citep{zhang2010nearly}. While MCP is often considered an improvement over SCAD \citep{zhang2010nearly,mazumder2011sparsenet,breheny2011coordinate}, extensive benchmarks \citep{bertsimas2016best, hastie2020best} have shown that the relaxed lasso yields performance comparable or superior to MCP across various settings. For these reasons, we focus on the relaxed lasso as the primary bias-corrected benchmark in our simulations, while further evaluating SCAD and MCP in the real-world data analysis.
%
\subsection{Sure Independence Screening (SIS)}
Sure Independence Screening (SIS) \citep{fan2008sure} is a simple and computationally efficient filter method that ranks predictors by their absolute marginal correlation with the response. Predictors are selected based on this ranking. Given a standardized predictor matrix $X$ and response vector $y$, this marginal correlation vector is $\rho_{xy}=X^\top y.$

Under certain regularity conditions, SIS enjoys the sure screening property, which ensures that the probability of the selected subset containing the true model approaches one as the number of observations tends to infinity \citep{fan2008sure}. However, SIS is limited by its reliance on marginal correlations. When predictors are correlated, marginal correlation rankings can misrepresent the true contribution of variables. For example, variables that are jointly important may appear irrelevant or weak when viewed marginally. This motivates the need for more comprehensive ranking measures, which we explore in the next section.
\section{Relative Importance Measures}\label{sec:relative importance}
This section introduces the concept of relative importance (RI) and presents a class of variable selection methods built upon RI measures. We begin with General Dominance (GD) and its practical approximations, then extend to high-dimensional generalizations. Finally, we formalize our rationale for using RI in the context of variable selection. This progression from theoretical definitions to computational implementation lays the necessary groundwork for the RI-based variable selection method and the proposed CRI.Z method.
\subsection{General Dominance (GD)}
In the presence of multicollinearity, simple measures such as marginal correlation or regression coefficient can yield misleading assessments of predictor importance. General Dominance (GD) \citep{budescu1993dominance,azen2003dominance} was developed to provide a more comprehensive evaluation. It defines the importance of a predictor as its average incremental contribution to the model fit---typically measured by the squared multiple correlation $R^2$---across all possible sub-models. For a predictor $x_i$, its GD is given by:
\begin{equation}
    \mathrm{GD}(x_i)=\frac{1}{p}\sum_{\mathcal{S}\subseteq \mathcal{P}\setminus\{i\}}\frac{1}{\binom{p-1}{|\mathcal{S}|}}\left(R_{y\cdot X_{\mathcal{S}\cup \{i\}}}^2-R_{y\cdot X_\mathcal{S}}^2\right),
    \label{eq:GD_definition}
\end{equation}
where $\mathcal{P}=\{1,...,p\}$, $\mathcal{S}\subseteq \mathcal{P}\setminus\{i\}$ denotes all possible subsets excluding the index of predictor $i$ and the term $R_{y\cdot X_{\mathcal{S}\cup \{i\}}}^2-R_{y\cdot X_\mathcal{S}}^2$ is the increase in $R^2$ from adding $x_i$ to a model containing the predictors in subset $\mathcal{S}$.

Conceptually, GD is equivalent to the Shapley value from cooperative game theory \cite{shapley1953game}, a principle now widely used in explainable AI \citep{lundberg2017unified}. By averaging over all sub-models, GD offers an equitable assessment of each predictor's contribution. However, it is computationally intractable for moderate to large $p$, as it requires fitting $2^p-1$ models.
\subsection{Relative Weight (RW) and Comprehensive Relative Importance (CRI)}
To address the computational challenge of GD, efficient approximations have been developed. Relative Weight (RW) \citep{genizi1993decomposition,johnson2000heuristic} is a widely used alternative that closely approximates GD \citep{leBreton2004monte,chao2008quantifying}. RW proceeds in three steps, collectively referred to as Orthogonalization-Reallocation Measures (ORMs) \cite{chang2025understanding}: it first transforms the correlated predictors $X$ into a set of maximally correlated orthogonal predictors $Z$ via the minimal transformation $Z=X(X^\top X)^{-1/2}$ \cite{johnson1966minimal}. The total explained variance of $y$ is allocated to $Z$ and then reallocated back to the original predictors based on the squared correlations between $x_i$ and $z_j$. While RW is more computationally efficient than GD, RW requires that the predictor matrix $X$ to be of full column rank.  In high-dimensional settings $(p>n)$ or when $X$ is singular, $(X^\top X)^{-1/2}$ is undefined. 

Comprehensive Relative Importance (CRI) \citep{shen2020comprehensive} overcomes this by generalizing RW to arbitrary $X$. It is derived using the reduced singular value decomposition (SVD) of predictor matrix, $X=U_rS_rV_r^\top,$
where $r$ is the rank of $X$, $U_r\in\mathbb{R}^{n\times r},S_r\in\mathbb{R}^{r\times r}$, and $V_r\in\mathbb{R}^{p\times r}$ are the first $r$ left singular vectors (column space of $X$), singular values, and right singular vectors (row space of $X$), respectively. CRI first constructs a generalized predictor matrix $Z_G$ that maximally correlated to $X$ (Supplementary Material Section A.2) and reallocates contributions of $Z_G$ back to $X$. The final CRI vector is then computed as:
\begin{equation}
    \label{eq:cri}
    D(X)=\underbrace{\left((V_rS_rV_r^\top)\odot(V_rS_rV_r^\top)\right)}_{\text{Contribution Reallocation}}\underbrace{\left((V_rU_r^\top y)\odot(V_rU_r^\top y)\right)}_{\text{Contribution of $Z_G$}},
\end{equation}
where $\odot$ denotes the Hadamard (element-wise) product. The second term allocates explained variance to $Z_G$, and the first term is the reallocation of the $Z_G$ contributions. When $X$ is full rank, CRI reduces to RW, making it a general tool for computing the relative importance. Since CRI is a generalized form of RW, we refer to both RW and CRI simply as CRI in what follows.
\subsection{An Alternative Importance Measure: CRI.Z}\label{subsec: CAR and CRI.Z}
An alternative importance measure can be derived by modifying the reallocation step in Eq.~\eqref{eq:cri}. By replacing the reallocation matrix (the first term) with an identity matrix, we obtain a simpler importance measure we term CRI.Z: 
\begin{equation}
    \label{eq:CRI.Z}
    w^2_G=(V_rU_r^\top y)\odot (V_rU_r^\top y).
\end{equation}
Since the reallocation term is an identity matrix in Eq.~\eqref{eq:CRI.Z}, the contributions of generalized orthogonal predictors to explain $y$ are assigned directly as the relative importance of the original predictors. 

When $n>p$, i.e. a full column rank of $X$, Eq.~\eqref{eq:CRI.Z} reduces to the squared marginal correlation between each orthogonal predictor and the response. This yields a vector of importance scores:
\begin{equation}
    \label{eq:CAR}
    w^2=[w_1^2,...,w_p^2]^\top,\,\text{where }w=
Z^\top y=(X^\top X)^{-1/2}X^\top y.
\end{equation}
This becomes the RI measure first introduced by \cite{johnson1966minimal} and later independently rediscovered in the variable selection literature as the Correlation-Adjusted marginal coRrelation (CAR) score \citep{zuber2011high}. The CAR score is based on a completely different concept inspired by decomposing the component in linear discriminant analysis \cite{zuber2009gene}.

The critical distinction between CAR and the proposed CRI.Z lies not only in their fundamentally different conceptual constructions but also in their treatment of covariance singularity (e.g., $p>n$). CAR relies on a heuristic shrinkage estimator (e.g., James–Stein type: $\lambda^\star I_p+(1-\lambda^\star)X^\top X$) to invert the singular covariance matrix \citep{schafer2005shrinkage}, which estimates the shrinkage parameter $\lambda^\star$ based on assumptions that may not be practical. In contrast, CRI.Z in Eq.~\eqref{eq:CRI.Z} is a completely different formulation that leverages the ORM framework used by CRI to provide a rigorous, parameter-free generalization. Thus, while CRI.Z and CAR are equivalent in low dimensions, CRI.Z offers a more theoretically consistent and principled solution for $p>n$ settings, avoiding the potential estimation bias or computational cost associated with heuristic shrinkage.

A comparison of the discussed RI measures is presented in Table \ref{tab:summary of RI measures}. In $n>p$ settings, it can be easily shown that CRI, CAR and CRI.Z share the same complexity $\mathcal{O}(np^2)$. However, in $p>n$ settings, CRI.Z is marginally faster than CAR by avoiding shrinkage parameter estimation and significantly faster than CRI by bypassing the construction of large $p\times p$ reallocation matrices. Nonetheless, for datasets with an exceptionally large $n$ or $p$, the $\mathcal{O}(\min(np^2,n^2 p))$ complexity of these RI-based methods makes them significantly more computationally demanding than marginal correlation (SIS), which scales linearly as $\mathcal{O}(np)$. A detailed computational complexity analysis is provided in Section A of the Supplementary Material.

\begin{table}[t]
    \centering
    \caption{Summary of the Relative Importance (RI) measures.}
    \label{tab:summary of RI measures}
    \begin{tabular}{lcp{2.2cm}p{2.7cm}}\hline
    Measure & Time Complexity $(p > n)$ & Handling $p>n$ & Core Idea  \\ \hline 
    GD \cite{budescu1993dominance} & $\mathcal{O}\left((np^2+p^3)2^p\right)$ & None & Shapley value \\ [0.5ex]
    CRI \cite{shen2020comprehensive} & $\mathcal{O}(n^2p+np^2)$ & Generalized predictors $Z_G$ & ORM \\ [0.5ex]
    CAR \cite{zuber2011high} & $\mathcal{O}(n^2p)$ & Shrinkage & Adjusted marginal correlations \\ [0.5ex]
    CRI.Z (ours) & $\mathcal{O}(n^2p)$ & Generalized predictors $Z_G$ & ORM \\ \hline
    \end{tabular}
\end{table}

\subsection{The RI-based Variable Selection Methods} 
Relative importance measures have traditionally been viewed as post-hoc explanatory tools. However, we offer a new perspective for GD or its close approximations to serve as an indicator for variable ranking and selection. This new role comes as no surprise if we look at the objective of best subset selection (Eq.~\eqref{eq:bs_prob}) that seeks the subset of predictors that maximizes the model fit (i.e., $R^2$). GD computes the average incremental contribution of each predictor to the $R^2$ by considering all possible sub-models. A predictor with a high GD contributes significantly, regardless of which other variables are in the model, to the $R^2$, suggesting that it is a strong candidate for inclusion.

Thus, while the best subset selection asks ``Which subset performs best?'', the RI measures ask ``Which predictors are most valuable to include in the subset?''. In other words, an indicator-based ranking and selection heuristic can be naturally developed based on the RI measures to approach the best subset problem. We now formalize our approach, which we term RI-based variable selection. This method falls into the class of filter methods. First, compute a ranking of all predictors using a chosen RI measure. Second, build a sequence of models by incrementally including predictors according to this ranking. 

This decouples the variable ranking from the model fitting. After computing importance scores, we fit models using the least squares method (LS-RI) or ridge regression (Ridge-RI) \citep{hoerl1970ridge} with the variables included based on their importance ranking. Ridge regression with regularization $\ell_2$ is intended to further improve the stability of the model \citep{shen2020comprehensive}. We do not use the lasso for the model fitting because its $\ell_1$ penalty performs a secondary variable selection step, which confounds the results of our primary selection method. The general RI-based selection algorithm is outlined in Algorithm 1 in Supplementary Material Section C.
\section{Simulations}\label{sec:simulation}
Our empirical evaluation of the proposed RI-based methods is organized in two parts. In Part I, we focus on the core task of variable ranking. Using the challenging simulation scenarios from Fan and Lv \cite{fan2008sure}, we evaluate the robustness of RI-based ranking (GD, CRI, CAR, CRI.Z) against the marginal correlation used by Sure Independence Screening (SIS). In Part II, we assess the predictive and selection performance of the RI-based models. For this, we adopt the comprehensive simulation framework from Hastie et al. \cite{hastie2020best}, enabling a rigorous comparison against established linear-model benchmarks, including best subset, forward stepwise, the lasso and relaxed lasso, across various levels of dimensionality, predictor correlation and Signal-to-Noise Ratio (SNR). The two-part simulation design allows us to separately evaluate the ranking capability of the measures (Part I) and their effectiveness in variable selection for model construction (Part II).

We focus our comparisons on these widely used benchmark methods because they represent the most influential and well-understood approaches across the wrapper, embedded, and filter methods. In line with this focus on representative benchmarks, we omit SCAD and MCP from our simulations; as discussed in Section \ref{subsec:lasso and relatives}, the relaxed lasso serves as a robust and highly competitive representative for bias-corrected regularization, having been shown to match or exceed the performance of these non-convex penalties in the simulation frameworks we adopt \citep{bertsimas2016best, hastie2020best}. While many recent methods based on meta-heuristic optimization or ensemble learning have shown promising results in specific applications \citep{de2020binary, meenachi2021metaheuristic, gao2025quantum, deng2013gene, zou2026feature}, these methods generally fall within the wrapper or embedded categories. We therefore do not include them in our evaluation, as our focus is on general-purpose baseline methods that can serve as a foundation upon which meta-heuristic or machine-learning–based strategies may be further developed.
\subsection{General Setup}
All simulations were conducted in \texttt{R} with the fixed random seed 42 for reproducibility. The data generation is based on the linear model $y=X\beta_0+\epsilon$. For each run, we construct a ground-truth coefficient vector $\beta_0\in\mathbb{R}^p$ with $s$ non-zero elements. The rows of the predictor matrix $X\in\mathbb{R}^{n\times p}$ are then drawn independently from $N_p(0,\Sigma)$ where $\Sigma=(\sigma_{ij})_{p\times p}$. Noise vector $\epsilon$ is drawn from $N_n(0,\sigma^2I)$, with variance 
$\sigma^2$ chosen to achieve a target $\text{SNR}=\beta_0^\top\Sigma\beta_0/\sigma^2$. 

Following \cite{hastie2020best}, we study four problem dimensions: low $(n=100,p=10)$, medium $(n=500,p=100)$, high-50 $(n=50,p=1000)$ and high-100 $(n=100,p=1000)$. Within each dimensions, we systematically vary predictor correlation and the SNR to evaluate performance across conditions. All simulation settings are summarized in Table \ref{tab:sim_summary}.

\begin{table}[t]
    \centering
    \caption{Summary of simulation settings.}
    \label{tab:sim_summary}
    \begin{tabular}{ll p{4cm}}
        \hline
        {Setting} & {Part I: Variable Ranking} & {Part II: Modeling} \\ \hline
        {$(n,p)$} & low, medium, high-50/100 & low, medium, high-50/100\\ 
        {$\rho$} & $\{0.35, 0.7, 0.9\}$ & $\{0, 0.35, 0.7, 0.9\}$ \\ 
        {SNR} & 4 levels: $\{0.05, 0.25, 1.22, 6\}$ & 10 levels: $\{0.05,\cdots, 6\}]$ \\ 
        {$\beta_0$} & beta-type 1, 2, 3 & beta-type 4, 5, 6 \\
        {Benchmark} & SIS & Best subset, forward stepwise, lasso, relaxed lasso, LS/Ridge-SIS \\ 
        {RI-based} & GD, CRI, CAR, CRI.Z & LS/Ridge-RI variants \\ \hline
    \end{tabular}%
\end{table}
\subsection{Part I: Variable Ranking}\label{sec:simulation part I}
The first set of simulation studies evaluate ranking robustness under scenarios known to be challenging for marginal correlation. 
\subsubsection{Setup for Ranking Comparison}
\paragraph{Competing methods} We compare the following variable ranking methods: (a) SIS, (b) GD, (c) CRI, (d) CAR, (e) Our proposed CRI.Z.

GD is only computed for the low dimension setting because its computational cost becomes too expensive to compute practically for the medium and high dimension settings. It is also important to note that in the $n>p$ settings, i.e., low and medium dimensions, CAR and CRI.Z are theoretically equivalent, despite being implemented through different approaches. We use the \texttt{R} packages \texttt{relaimpo} \cite{gromping2007relative} and \texttt{care} \cite{zuber2021care} for the implementation of GD and CAR, respectively, while the authors implement CRI and CRI.Z directly in \texttt{R}.
\paragraph{Simulation beta-types} We consider the three challenging cases from \citep{fan2008sure}:
\begin{itemize}[noitemsep, topsep=0pt]
    \item \textit{beta-type 1.} Equicorrelated predictors ($\sigma_{ij}=\rho,\,\forall i\neq j$ and $\sigma_{ii}=1,\,\forall i=1,...,p$), with $s=3$ strong signals: $\beta_0=[5_{s\times 1}^\top,0_{(p-s)\times1}^\top]^\top$. 

    \item \textit{beta-type 2.} 
    Extending beta-type 1 with an additional suppressor variable $x_4$ (thus $s=4$), yielding $\beta_0=[5_{(s-1)\times 1}^\top,-15\rho^{1/2},0^\top_{(p-s)\times1}]^\top$. $x_4$ has correlation $\rho^{1/2}$ with all other $p-1$ variables but has zero marginal correlation with the response.
    
    \item \textit{beta-type 3.} Extending beta-type 2 by adding a weak predictor $x_5$ (thus $s=5$), yielding $\beta_0=[5_{(s-2)\times 1}^\top,-15\rho^{1/2},1,0_{(p-s)\times1}^\top]^\top$. $x_5$ is uncorrelated with all other $p-1$ variables but has a weak marginal correlation.
\end{itemize}
We consider three predictor correlation levels $\rho\in\{0.35,0.7,0.9\}$ and four SNR values $\text{SNR}=\{0.05,0.25,1.22,6\}$. 
\paragraph{Evaluation metrics} Given a variable ranking, we evaluate its performance using two criteria:
\begin{itemize}[noitemsep, topsep=0pt]
    \item $S$: the minimal model size required to include all true predictors.
    \item ${\Pr}(k)$: the proportion of true predictors among the top-$k$ selected variables. 
\end{itemize}
All results are averaged over 100 replications. For the low dimension setting, we set $k$ up to 10 and for all other settings, we set $k$ up to 50.
\subsubsection{Results for Ranking Performance}
\begin{figure}[htb!]
    \centering
    \includegraphics[width=1\linewidth]{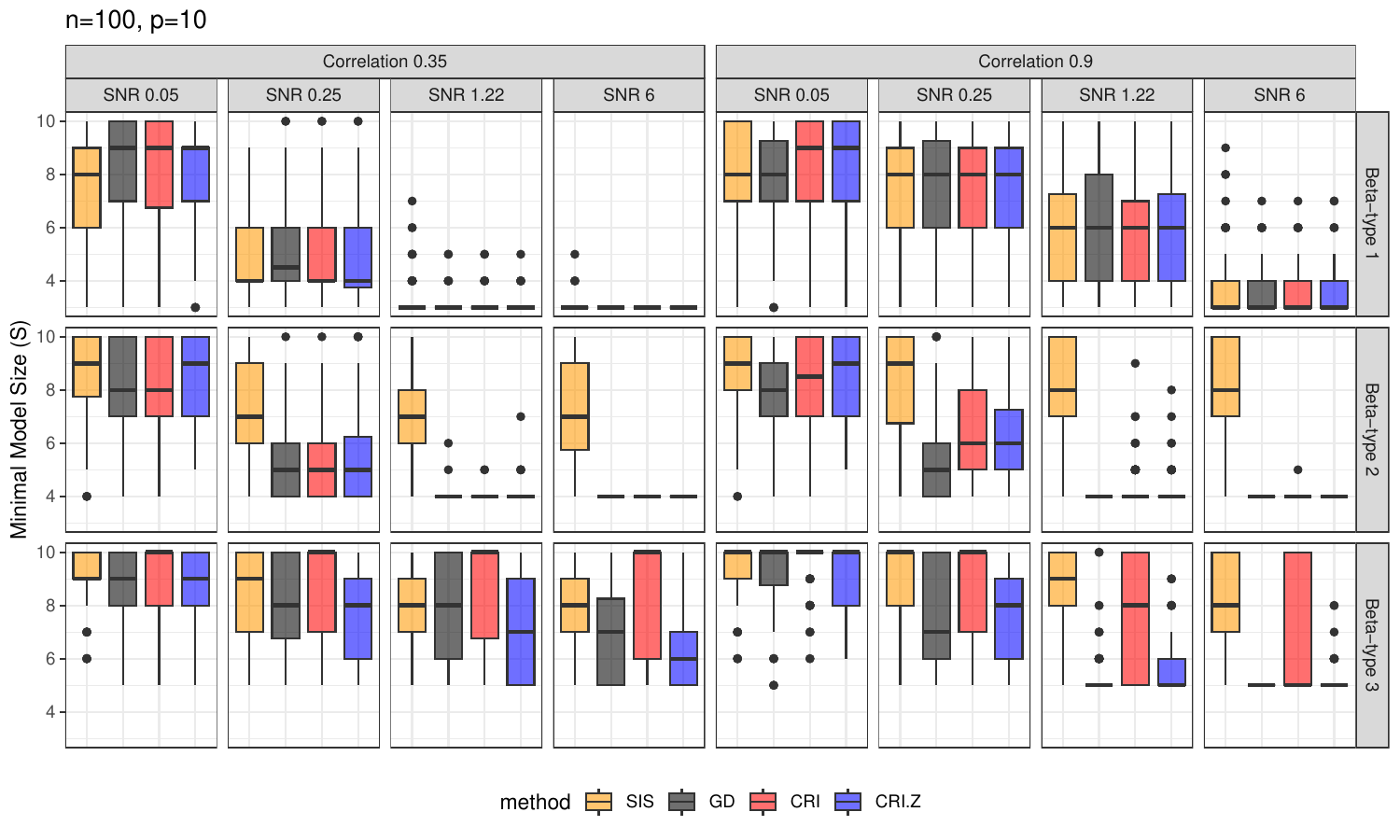}
    \caption{Boxplots for $S$ for the GD, CRI, CRI.Z, and SIS methods for $\rho\in\{0.35,0.7,0.9\}$ and $\text{SNR}\in\{0.05,0.25,1.22,6\}$ based on 100 replications under different beta-types with $(n,p)=(100,10)$.}
    \label{fig:low S}
\end{figure}

\begin{figure}[htb!]
    \centering
    \includegraphics[width=1\linewidth]{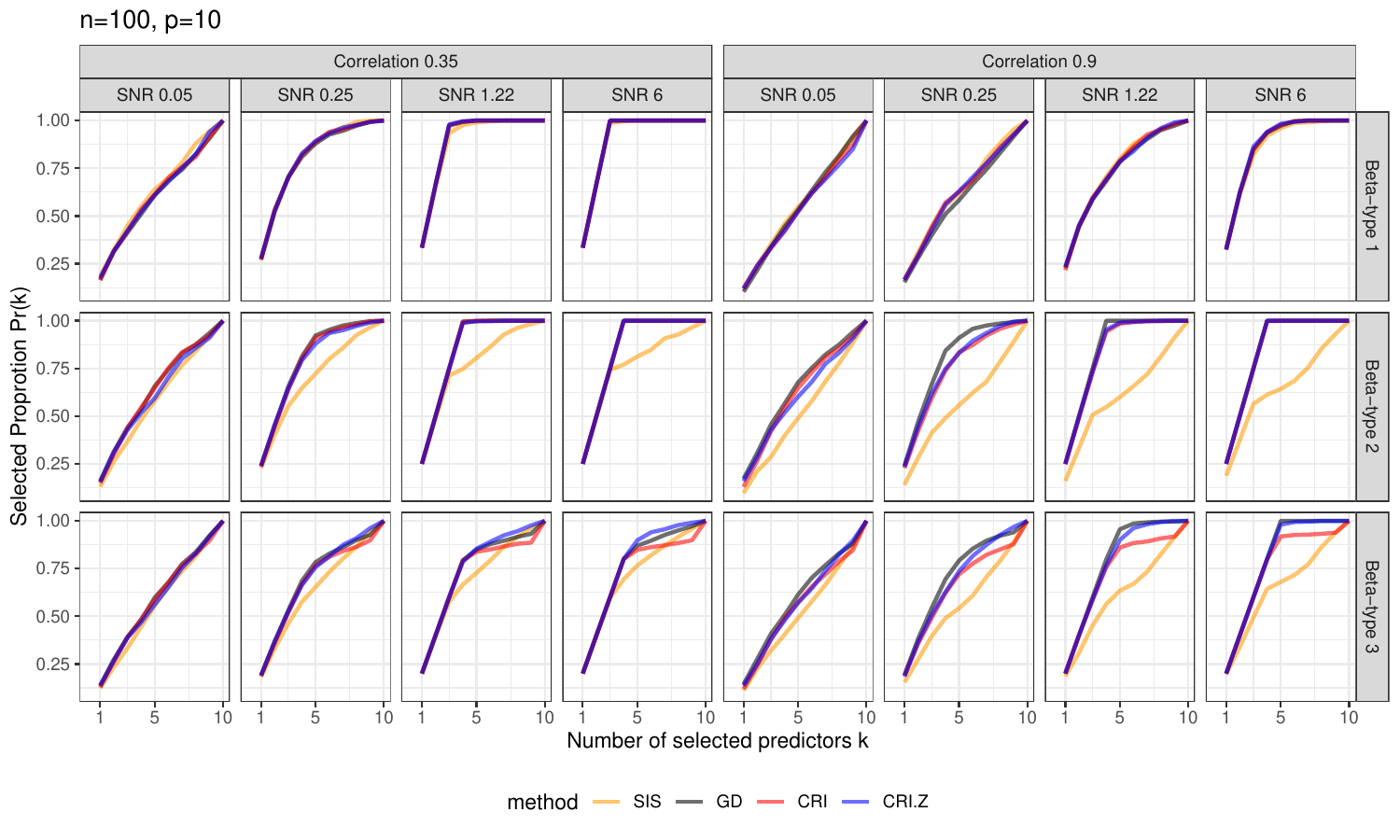}
    \caption{Summary results for $\Pr(k)$ for the GD, CRI, CRI.Z, and SIS methods for $\rho\in\{0.35,0.7,0.9\}$ and $\text{SNR}\in\{0.05,0.25,1.22,6\}$ based on 100 replications under different beta-types with $(n,p)=(100,10)$.}
    \label{fig:low Pr}
\end{figure}
We present results for the low ($n=100,p=10$) and high-100 ($n=100,p=1000$) dimension settings in Figs. \ref{fig:low S} to \ref{fig:hi10 Pr}. Full results are available in the Section B of Supplementary Material.

In the low dimension setting (Figs. \ref{fig:low S}--\ref{fig:low Pr}), all methods perform well in beta-type 1, though the performance degrades as the correlation increases. In the harder beta-types 2 and 3, SIS fails due to its reliance on marginal correlations, whereas RI measures (GD, CRI, CRI.Z) remain robust. Subtle differences emerge: GD and CRI slightly outperform CRI.Z in beta-type 2 with a suppressor, while CRI.Z performs better than CRI in beta-type 3 with an extra weak predictor, suggesting its simple reallocation after orthonormal transformation is advantageous for weak signals.

\begin{figure}[tb]
    \centering
    \includegraphics[width=1\linewidth]{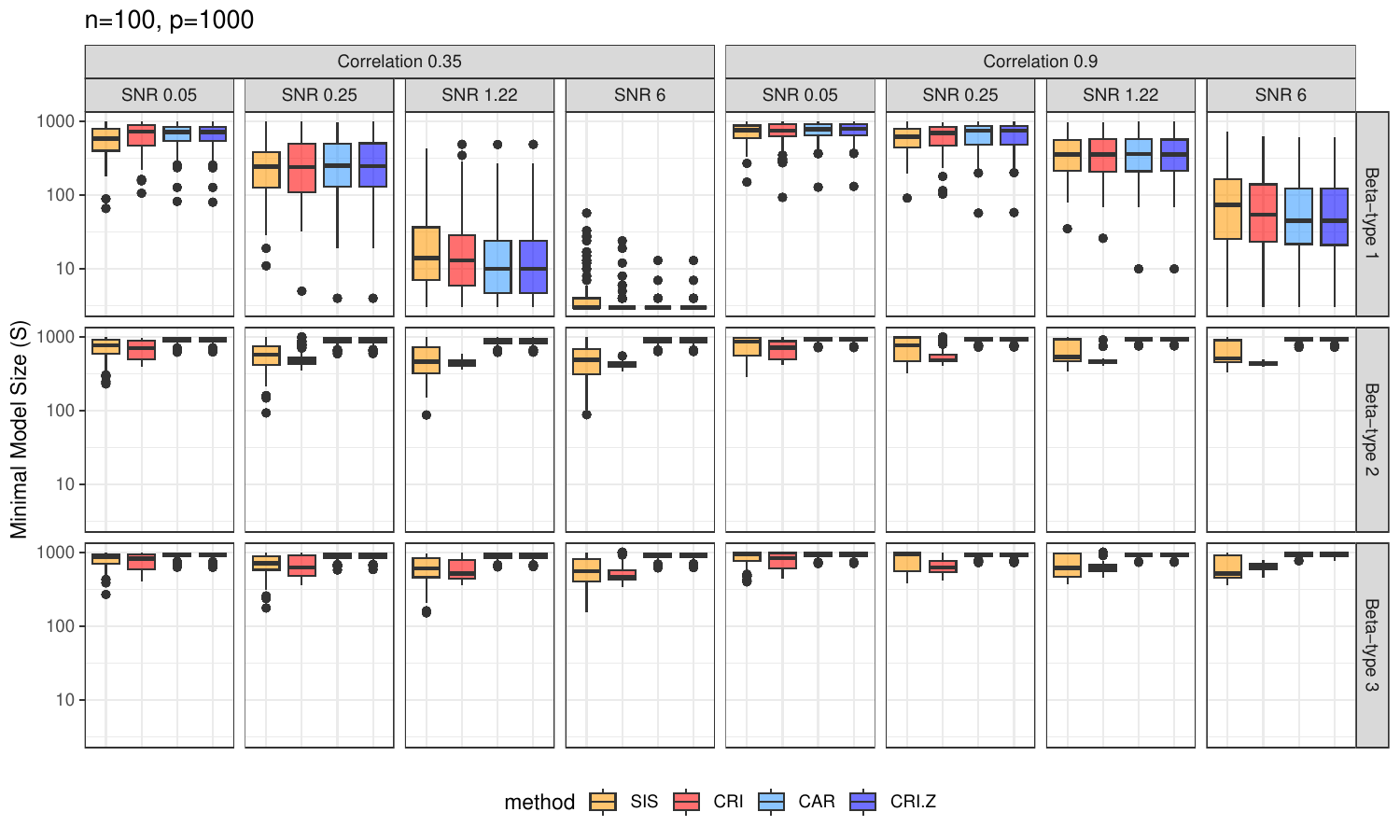}
    \caption{Boxplots for $S$ for the GD, CRI, CRI.Z, and SIS methods for $\rho\in\{0.35,0.7,0.9\}$ and $\text{SNR}\in\{0.05,0.25,1.22,6\}$ based on 100 replications under different beta-types with $(n,p)=(100,1000)$.}
    \label{fig:hi10 S}
\end{figure}

\begin{figure}[tb]
    \centering
    \includegraphics[width=1\linewidth]{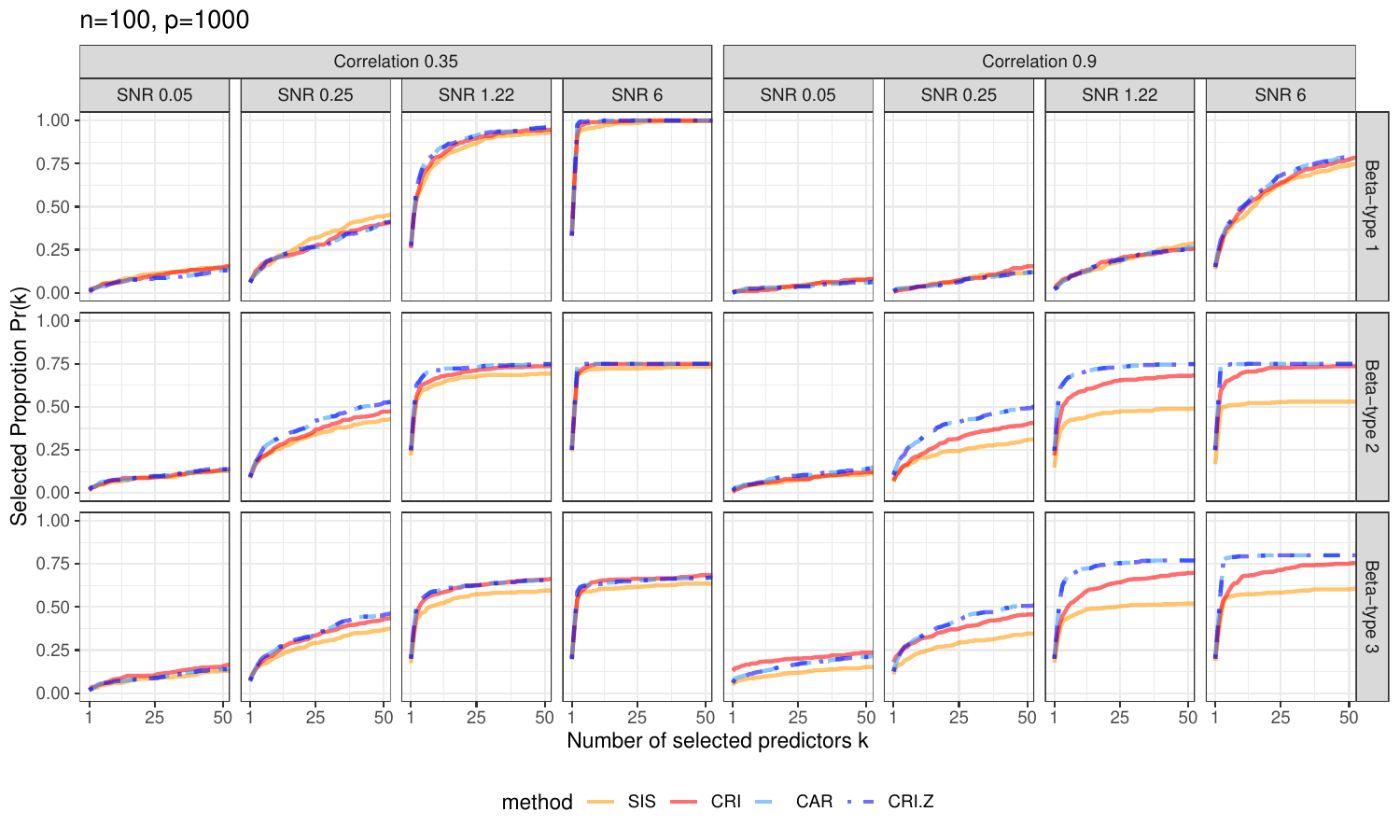}
    \caption{Summary results for $\Pr(k)$ for the CRI, CAR, CRI.Z, and SIS methods for $\rho\in\{0.35,0.7,0.9\}$ and $\text{SNR}\in\{0.05,0.25,1.22,6\}$ based on 100 replications under different beta-types with $(n,p)=(100,1000)$.}
    \label{fig:hi10 Pr}
\end{figure}

In the high-100 dimension setting (Figs. \ref{fig:hi10 S}--\ref{fig:hi10 Pr}), SIS again performs poorly in beta-types 2 and 3. Among RI measures, CRI.Z and CAR consistently outperform CRI, particularly under higher correlations. That highlights that a simple identity reallocation from the orthogonal predictors to original predictors is quite effective in high dimensions. It should be noted, however, although that CRI.Z and CAR attain higher selection proportions $\Pr(k)$, none of the methods can perfectly recover the true predictors, especially those suppressed or weak compared to noise, as reflected by the large $S$ required to capture true predictors in Fig. \ref{fig:hi10 S} for beta-type 2 and beta-type 3.

In summary, RI measures offer a more reliable ranking than marginal correlation, especially in the presence of suppressors or weak predictors. This motivates their use as filter-based selection methods, examined in Part II.
\subsection{Part II: Modeling}\label{sec:simulation part II}
We now evaluate complete RI-based selection methods, comparing them against benchmarks in both support recovery and predictive accuracy.
\subsubsection{Setup for Modeling Comparison}
\paragraph{Competing methods} We compare the following variable selection methods: (a) best subset, (b) forward stepwise, (c) lasso, (d) relaxed lasso, (e) LS-SIS (using SIS ranking with least squares fit) and (f) LS-RI variants where RI are one of $\{\text{GD, CRI, CAR and CRI.Z}\}$. 

Benchmark methods were implemented using the public repository of Hastie et al. \cite{hastie2020best}. Due to the impractically expensive computation cost, best subset and our LS-GD were only evaluated in the low dimension setting. While our primary analysis focuses on the LS-RI variants, also included is the regularized Ridge-CRI.Z in the low dimension plots (Figs. \ref{fig:low F1} and \ref{fig:low error}) to illustrate the benefits of regularization. Extended evaluation of all Ridge-RI variants and corresponding $\ell_2$ penalty tuning, comparable to the relaxation tuning in the relaxed lasso, is provided in Supplementary Section C.
\paragraph{Simulation beta-types} We consider three beta-types from \citep{hastie2020best}: 
\begin{itemize}[noitemsep, topsep=0pt]
    \item \textit{beta-type 4.} All $s$ non-zero coefficients are set to 1 and evenly spaced.
    \item \textit{beta-type 5.} $\beta_0=[1_{s\times1}^\top,0_{(p-s)\times 1}^\top]^\top$.
    \item \textit{beta-type 6.} $\beta_{0,i}=0.5+(10-0.5)\frac{(i-1)}{s-1},\,\forall i=1,...,s$ and $\beta_{0,i}=0,\,\forall i=s+1,...,p$.
\end{itemize}
The predictor covariance matrix is set to be $\sigma_{ij}=\rho^{|i-j|},\,i,j=1,...,p$. We consider four predictor correlation levels $\rho\in\{0,0.35,0.7,0.9\}$ and ten SNR values $\text{SNR}\in\{0.05,...,6\}$ following the setup in \cite{hastie2020best}. For low, medium and high-50 dimension settings, we set $s$ to 5 and for high-100 setting, we set $s$ to 10.
\paragraph{Evaluation metrics} We evaluate the performance of each method using two key metrics adopted from \cite{hastie2020best} given an estimated coefficients $\hat{\beta}$ from one of the methods.
\begin{itemize}[noitemsep, topsep=0pt]
     \item \textit{F1-score}: Accuracy of support recovery, ranging from 0 to 1.
     \item \textit{Relative Test Error (RTE)}: 
     $\text{RTE}(\hat{\beta})={(\hat\beta-\beta_0)^\top\Sigma(\hat\beta-\beta_0)}/{\sigma^2}.$
\end{itemize}
All results are averaged over 30 replications. 
\paragraph{Tuning procedures} In all cases, tuning was performed by minimizing prediction error on an external validation set of size $n$, which is independently and identically generated as in \cite{hastie2020best}. The tuning parameters for benchmark methods are also set as in \cite{hastie2020best}. The only parameter to be tuned for the LS-SIS and LS-RI methods is the number of variables to include in the model $k$, which is comparable to the $\ell_1$ penalty in the lasso and relaxed lasso methods. In the low dimension setting, $k$ is tuned over the range of $k=0,...,10$. In all other problem settings (medium, high-50, and high-100 dimensions), the parameter is tuned over $k=0,...,50$. For two-parameter Ridge-RI variants (tuning both $k$, and penalty $\lambda$), we perform a grid search to jointly identify the optimal parameter pair. Detailed specifications for the $\lambda$ grids are provided in Supplementary Material Section C.
\subsubsection{Results for Modeling Performance}
\begin{figure}[htb!]
    \centering
    \includegraphics[width=1\linewidth]{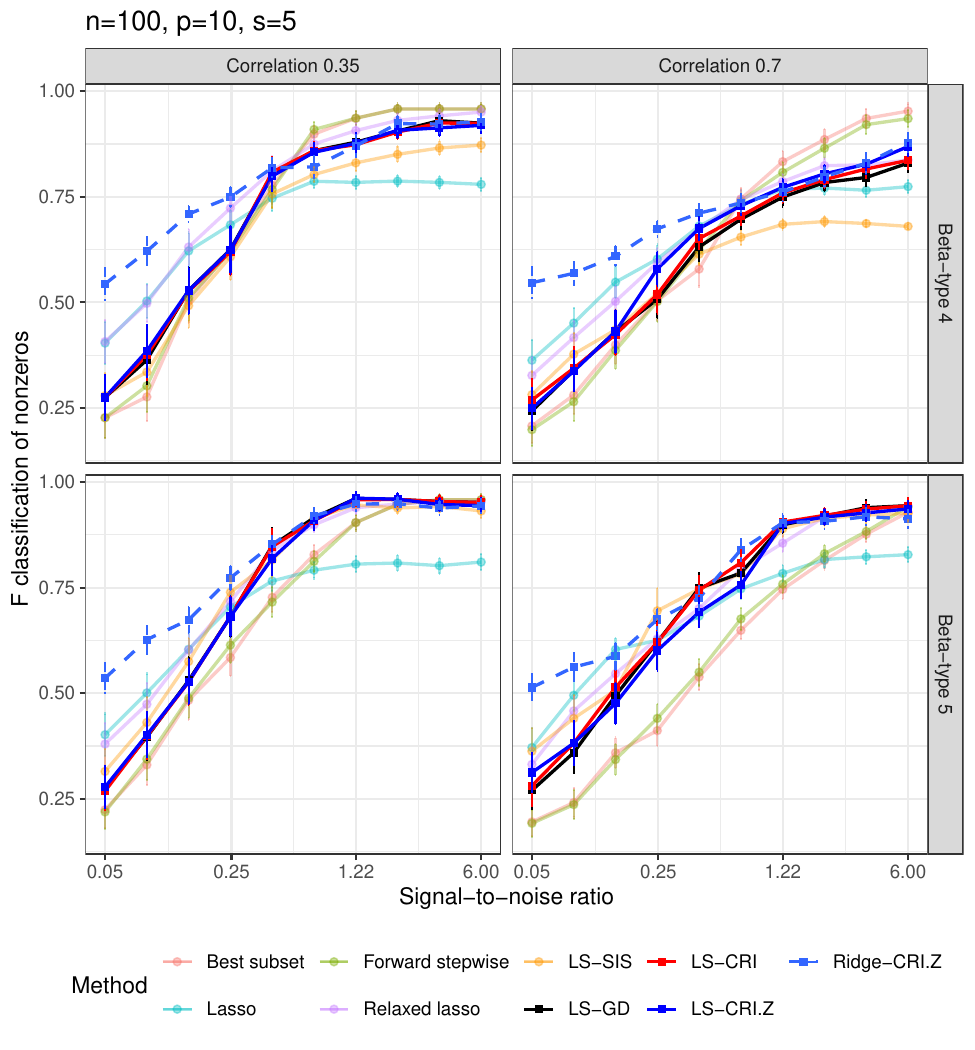}
    \caption{F1-score as function of SNR in the low setting with $n=100,p=10$ and $s=5$.}
    \label{fig:low F1}
\end{figure}

\begin{figure}[htb!]
    \centering
    \includegraphics[width=1\linewidth]{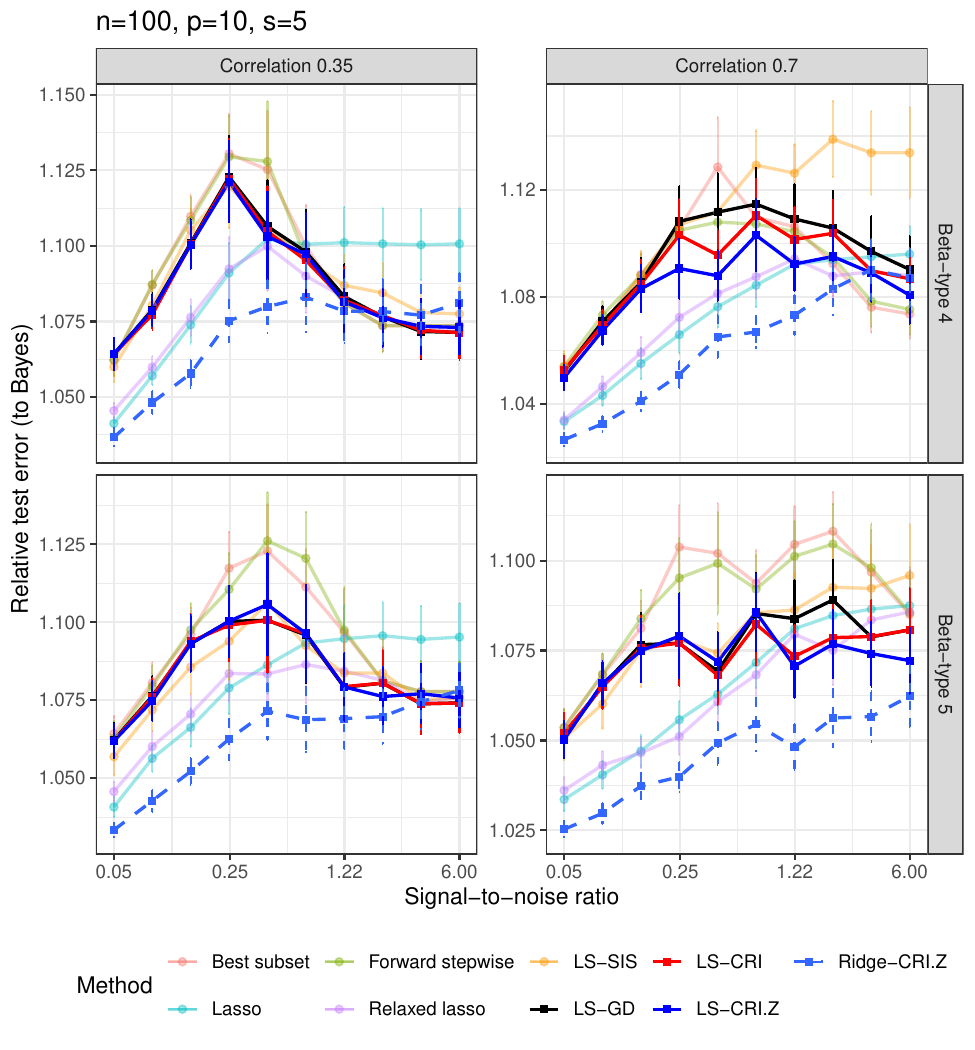}
    \caption{RTE as function of SNR in the low setting with $n=100,p=10$ and $s=5$.}
    \label{fig:low error}
\end{figure}

\begin{figure}[htb!]
    \centering
    \includegraphics[width=1\linewidth]{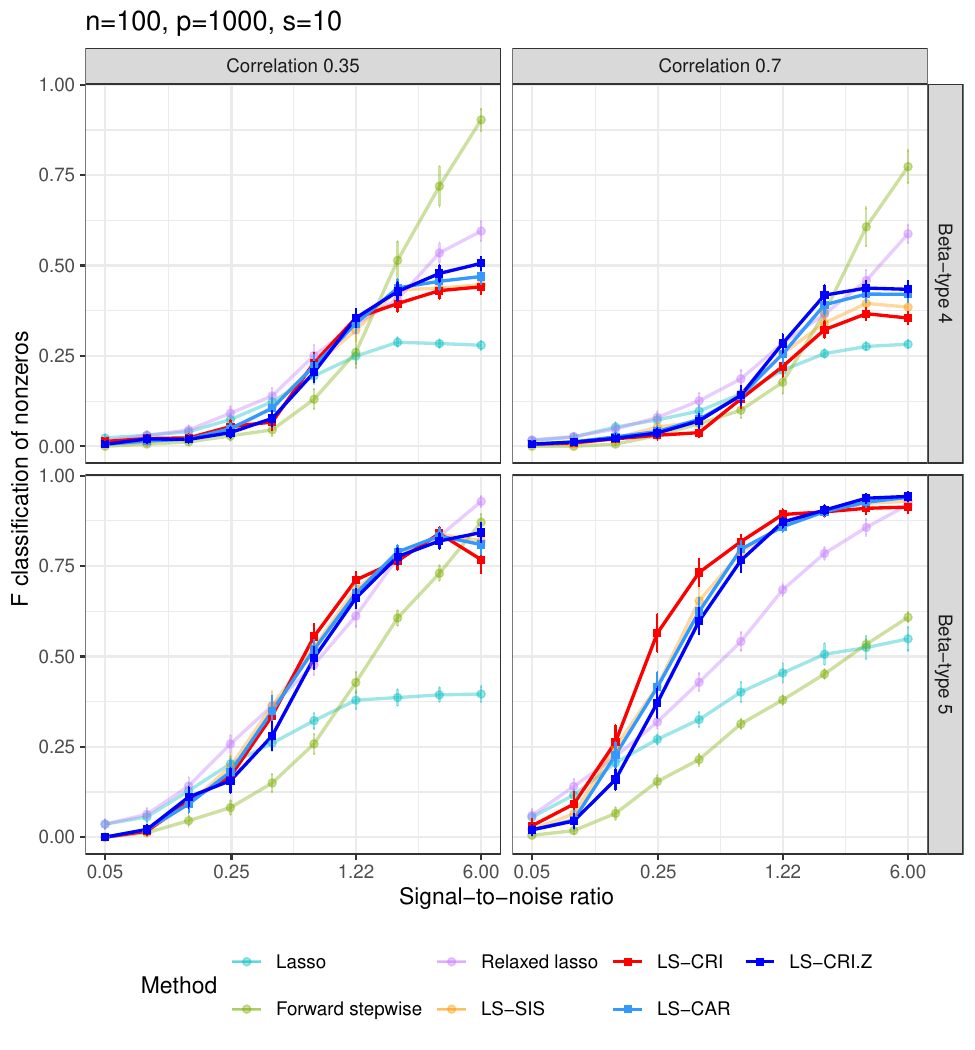}
    \caption{F1-score as function of SNR in the high-100 setting with $n=100,p=1000$ and $s=10$.}
    \label{fig:hi10 F1}
\end{figure}

\begin{figure}[htb!]
    \centering
    \includegraphics[width=1\linewidth]{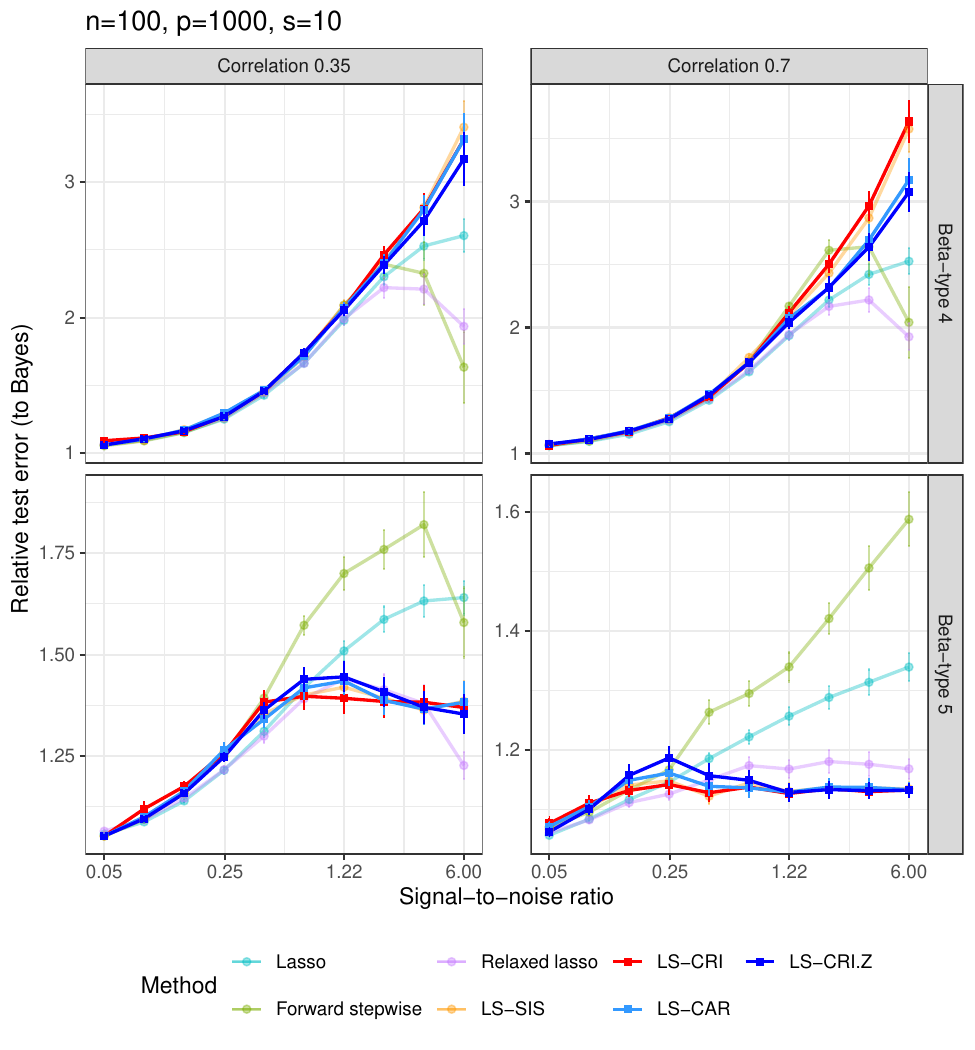}
    \caption{RTE as function of SNR in the high-100 setting with $n=100,p=1000$ and $s=10$.}
    \label{fig:hi10 error}
\end{figure}

We present results for the low ($n=100,p=10$) and high-100 ($n=100,p=1000$) dimension settings in Figs. \ref{fig:low F1} to \ref{fig:hi10 error}. Each panel displays the averaged metric across 30 repetitions, with vertical bars indicating one standard error. The following analyses focus on cases with correlation levels $\rho\in\{0.35,0.7\}$ in beta-types 4 and 5, as beta-type 6 yields similar conclusions to beta-type 5. Full results are provided in Supplementary Material Section C. 

In the low dimension setting (Figs. \ref{fig:low F1} and \ref{fig:low error}) the results for beta-type 4 (Fig. \ref{fig:low F1}, upper panel) align with the bias–variance trade-off discussed in \cite{hastie2020best}, with the lasso and relaxed lasso excelling at low SNR and best subset and forward stepwise performing better as SNR increases. In this context, the LS-RI variants strike an effective balance. Their performance typically falls between those of the lasso and the best subset. Among the RI-based methods, LS-GD and LS-CRI perform almost identically. LS-CRI.Z consistently outperforms LS-CRI and closely matches or even exceeds the relaxed lasso at a higher SNR, especially under stronger predictor correlation. LS-SIS, on the contrary, is more sensitive to the correlations among the predictors, causing it to underperform.

The advantages of RI-based selection are more pronounced in beta-type 5 (Fig. \ref{fig:low F1}, lower panel), which features a clustered predictor structure. Here, the performance of best subset and forward stepwise selection deteriorates sharply with increasing predictor correlations. Specifically, at SNR 1.22, the F1-score of these methods drops from approximately 0.9 to 0.75 as correlation increases from 0.35 to 0.7. In contrast, the LS-RI variants achieve superior support recovery, attaining the highest F1-scores across all but the lowest SNR levels. Their performance advantage over the relaxed lasso widens as the correlation increases, highlighting the robustness of RI-based methods in settings with highly correlated predictors.

Analysis of the RTE in Fig. \ref{fig:low error} reinforces these observations. Although the LS-RI variants achieve high F1-scores, both lasso and relaxed lasso yield lower RTE due to their variance reduction by $\ell_1$ regularization. However, this gap can be filled by applying the $\ell_2$ regularization to our method. The Ridge-CRI.Z method (dashed line) thus not only closes the RTE gap but outperforms the relaxed lasso on both F1-score and RTE.

In the high-100 dimension setting (Figs. \ref{fig:hi10 F1} and \ref{fig:hi10 error}), the patterns observed in beta-type 4 are consistent with the low dimension setting. The LS-RI methods again achieve a very good balance, with LS-CRI.Z and LS-CAR emerging as the best. In high-SNR cases, LS-CRI.Z tends to outperform LS-CAR, suggesting that CAR's inherent shrinkage may introduce estimation bias, similar to the lasso. This result confirms the scalability and robustness of the proposed framework. As before, the superiority of RI-based methods is most evident in the clustered predictor scenario of beta-type 5 (Fig. \ref{fig:hi10 F1}, lower panel). In this challenging setting, the performance of forward stepwise collapses under the high predictor correlation; for example, at SNR 6.00, the F1-score of forward stepwise drops from around 0.9 to 0.6 as correlation increases from 0.35 to 0.7. In contrast, the LS-RI variants consistently achieve the highest F1-scores. The performance gap over the relaxed lasso widens in high dimensions, underscoring both the stability and scalability of RI-based selection and modeling. The RTE results again show that this advantage is matched and often exceeded with Ridge-RI's $\ell_2$ regularization (Supplementary Material Section C).

In beta-type 5, LS-SIS matches the LS-RI variants in both low and high dimensions. This is because the clustered true predictors produce high marginal correlations, aligning with the conditions under which SIS is effective. This contrasts with its underperformance in beta-type 4 and its failure in the suppressed signal scenarios discussed previously. While SIS can perform well under favorable conditions, its sensitivity to data structure limits its general applicability. The RI-based methods, in contrast, demonstrate consistent and robust performance across all tested scenarios.

A detailed comparison of average computational runtimes is provided in Table S1 of Supplementary Material Section C. Among the benchmarks, best subset is the most computationally intensive, becoming intractable beyond low dimension setting. Forward stepwise selection, while avoiding exponential complexity, scales poorly with dimension and emerges as the slowest method in the medium and high dimension settings. Conversely, the lasso and relaxed lasso are highly efficient. However, this speed is largely attributable to mature algorithmic optimizations. While early LARS implementations scaled as $\mathcal{O}(p^3+np^2)$ \cite{efron2004least}, modern coordinate descent algorithms (e.g., \texttt{glmnet}) utilize optimized Fortran backends to achieve near-linear time complexity $\mathcal{O}(np)$ \cite{friedman2010regularization}.

Regarding the proposed RI-based methods, LS-CRI, LS-CAR, and LS-CRI.Z exhibit similar runtimes in $n>p$ settings. However, significant distinctions emerge in high dimensional settings. LS-CRI becomes computationally heavier due to the $\mathcal{O}(n^2p+np^2)$ cost of constructing large reallocation matrices. LS-CRI.Z proves to be the most efficient implementation, significantly outperforming LS-CRI and marginally outperforming LS-CAR, as it avoids both the complex matrix operations of CRI and the shrinkage parameter estimation required by CAR.

In summary, the proposed RI-based methods offer a robust and scalable alternative for variable selection and modeling. They are particularly effective in high correlation settings where existing methods often struggle. The LS-RI variants, especially LS-CRI.Z and its regularized version Ridge-CRI.Z, offer simple yet powerful solutions that outperform linear-model benchmarks in both selection accuracy and prediction error across rigorous simulations.
\section{Real-World Dataset Examples}\label{sec:real-world}
To illustrate the utility of the proposed methods in practical high-dimensional settings, we analyzed two benchmark gene expression datasets from the Arizona scikit-feature repository\footnote{\url{https://jundongl.github.io/scikit-feature/}} \citep{li2017feature}. In this section, we first detail the experimental setup and evaluation metrics, followed by a comparative analysis of classification accuracy, model size, and computational efficiency for the Leukemia and GLI\_85 datasets.

\subsection{Experimental Setup}
For each dataset, we adopted a rigorous repeated random splitting strategy. In each of 100 repetitions, the data was stratified and split into training (64\%), validation (16\%), and testing (20\%) sets. Since the response variable is binary, the linear regression model is equivalent to linear discriminant analysis \cite{nie2022equivalence}. Therefore, we fit a linear regression model, and classify observations as class 1 if the predicted value $> 0.5$, and class 0 otherwise.

Hyperparameters were tuned on the validation set via grid search. For forward stepwise, LS-SIS, and LS-RI variants, the model size was tuned over $k \in \{1, \dots, n\}$. For the lasso and relaxed lasso, we tuned the penalty $\lambda$ over a sequence of length $2n$. SCAD and MCP used the same $\lambda$ sequence, with the convexity parameters tuned via grid search (see Supplementary Material Section D). For Ridge-SIS and Ridge-RI variants, we jointly tuned $k \in \{1, \dots, n\}$ and $\lambda$ (20 values). We report the F1-score and Model Size (number of selected variables). Given the class imbalance in both datasets, we also report the Balanced Accuracy (Bal. Acc.), defined as the average of sensitivity and specificity.

\subsection{Leukemia Dataset}
The Leukemia dataset \cite{golub1999molecular} contains $p=7129$ gene expression levels and $n=72$ patients, comprising 47 acute lymphocytic leukemia (ALL) and 25 acute myelogenous leukemia (AML). The results are summarized in Table \ref{tab:leukemia}. Among the benchmarks, the relaxed lasso achieves the highest performance, while forward stepwise performs poorly due to its greedy nature. The proposed RI-based methods demonstrate distinct advantages. First, the performance of Ridge-CRI.Z matches the top-performing relaxed lasso while selecting a comparable number of variables. Furthermore, it outperforms Ridge-SIS while yielding a significantly smaller model size. Second, while the LS-RI variants (LS-CRI.Z, LS-CAR) show a slightly lower accuracy compared to the lasso, they produce significantly smaller models (selecting $\approx 15$ genes vs. 33 for the lasso). LS-CRI.Z yields performance comparable to SCAD and MCP but selects significantly fewer variables even than MCP, which is recognized for its sparse selection \citep{breheny2011coordinate}. This compactness can be highly valuable in genomic biomarker discovery. Finally, although LS-CAR and LS-CRI.Z yield statistically similar predictive performance, LS-CRI.Z is more efficient (14.9 ms vs. 18.3 ms). Both are orders of magnitude faster than the original LS-CRI (224.6 ms), SCAD (162.1 ms) and MCP (159.5 ms).

\begin{table}[tb]
    \centering
    \caption{Performance results for the Leukemia dataset \cite{golub1999molecular}.}
    \label{tab:leukemia}
    \begin{tabular}{lrrrr}
        \hline
        Method & Bal. Acc. & F1-score & Model Size & Time (ms) \\ \hline
        \textit{Baselines} & & & & \\
        Lasso & 0.92 (0.01) & 0.9 (0.01) & 33.38 (1.28) & 3.9 (0.36) \\ 
        Forward stepwise & 0.84 (0.01) & 0.77 (0.02) & 4.84 (0.61) & 112.0 (2.55) \\ 
        Relaxed lasso & \textbf{0.93} (0.01) & \textbf{0.91} (0.01) & 24.63 (1.33) & 4.2 (0.37) \\
        SCAD & 0.91 (0.01) & 0.88 (0.01) & 25.21 (0.82) & 162.1 (4.62) \\ 
        MCP & 0.91 (0.01) & 0.89 (0.01) & 19.37 (0.98) & 159.5 (4.30) \\ \hline
        \textit{LS-Based} & & & & \\
        LS-SIS & 0.86 (0.01) & 0.82 (0.01) & 15.75 (1.08) & 2.3 (0.27) \\ 
        LS-CRI & 0.86 (0.01) & 0.81 (0.01) & 13.51 (0.96) & 224.6 (4.62) \\ 
        LS-CAR & 0.90 (0.01) & 0.88 (0.01) & 15.17 (1.03) & 18.3 (0.85) \\
        LS-CRI.Z & 0.90 (0.01) & 0.88 (0.01) & 15.35 (1.04) & 14.9 (0.70) \\ \hline
        \textit{Ridge-Based} & & & & \\
        Ridge-SIS &0.91 (0.01) & 0.89 (0.01) & 32.02 (1.33) & 59.8 (1.81) \\ 
        Ridge-CRI & 0.90 (0.01) & 0.88 (0.01) & 30.37 (1.44) & 291.6 (5.40) \\ 
        Ridge-CAR &\textbf{0.93} (0.01) & \textbf{0.91} (0.01) & 25.34 (1.43) & 77.1 (1.83) \\
        Ridge-CRI.Z & \textbf{0.93} (0.01) & \textbf{0.91} (0.01) & 25.17 (1.40) & 68.9 (1.72) \\ 
        \hline
        \multicolumn{5}{l}{\footnotesize Notes: Values represent mean (standard error). Bal. Acc.: Average of sensitivity and specificity.}
    \end{tabular}
\end{table}

\subsection{GLI\_85 Dataset}
For the second dataset, we selected GLI\_85 dataset \cite{freije2004gene} that has a much larger number of predictors ($p=22,283$) than the Leukemia dataset while the sample size ($n=85$ gliomas) is not much greater, with 26 Grade III and 59 Grade IV. Table \ref{tab:GLI_85} presents the results. In this extremely high-dimensional setting, the proposed Ridge-CRI.Z achieves the best overall performance, outperforming both the lasso and relaxed lasso. Similar to the Leukemia results, the LS-RI variants select significantly fewer variables than the benchmarks (model size $\approx 16$ vs. 36 for lasso, 29 for SCAD and 19 for MCP) while maintaining competitive accuracy. Notably, the computational advantage of CRI.Z is pronounced in this setting. LS-CRI.Z (53.5 ms) is orders of magnitude faster than LS-CRI (2411.4 ms) and faster than LS-CAR (69.8 ms) and over ten times faster than SCAD (628.0 ms) and MCP (612.9 ms). This confirms that CRI.Z successfully scales to extremely high predictor-to-sample ratio ($p/n>250$) while achieving superior predictive power.
\begin{table}[tb]
    \centering
    \caption{Performance results for the GLI\_85 dataset \cite{freije2004gene}.}
    \label{tab:GLI_85}
    \begin{tabular}{lrrrr}
        \hline
        Method & Bal. Acc. & F1-score & Model Size & Time (ms) \\ \hline
        \textit{Baselines} & & & & \\
        Lasso & 0.79 (0.01) & \textbf{0.90} (0.00) & 36.03 (1.78) & 16.6 (0.85) \\ 
        Forward stepwise & 0.75 (0.01) & 0.86 (0.01) & 3.72 (0.51) & 519.0 (8.88) \\ 
        Relaxed lasso & 0.78 (0.01) & 0.88 (0.01) & 21.97 (1.85) & 10.4 (0.48) \\
        SCAD & 0.79 (0.01) & 0.89 (0.01) & 29.76 (1.11) & 628.0 (16.34) \\ 
        MCP & 0.76 (0.01) & 0.87 (0.01) & 19.02 (1.36) & 612.9 (15.99) \\ \hline
        \textit{LS-Based} & & & & \\
        LS-SIS & 0.79 (0.01) & 0.88 (0.01) & 9.85 (0.87) & 9.0 (0.53) \\ 
        LS-CRI & 0.78 (0.01) & 0.89 (0.01) & 13.86 (1.10) & 2411.4 (36.1) \\ 
        LS-CAR & 0.77 (0.01) & 0.87 (0.01) & 16.78 (1.17) & 69.8 (1.82) \\ 
        LS-CRI.Z & 0.77 (0.01) & 0.87 (0.01) & 16.80 (1.17) & 53.5 (1.60) \\ \hline
        \textit{Ridge-Based} & & & & \\
        Ridge-SIS & \textbf{0.81} (0.01) & \textbf{0.90} (0.01) & 23.47 (1.76) & 197.8 (3.59) \\ 
        Ridge-CRI & \textbf{0.81} (0.01) & \textbf{0.90} (0.01) & 30.86 (1.61) & 2828.8 (42.2) \\ 
        Ridge-CAR & \textbf{0.81} (0.01) & \textbf{0.90} (0.00) & 31.45 (1.68) & 250.1 (4.96) \\ 
        Ridge-CRI.Z & \textbf{0.81} (0.01) & \textbf{0.90} (0.00) & 31.3 (1.68) & 235.3 (4.41) \\ \hline
        \multicolumn{5}{l}{\footnotesize Notes: Values represent mean (standard error).}
    \end{tabular}
\end{table}

\section{Discussion}\label{sec:discussion}
This study establishes that variable selection methods based on relative importance (RI) rankings are a robust and competitive alternative to traditional approaches. This section analyzes the proposed methods through the lens of model complexity, investigates the performance differences among RI measures, discusses potential limitations and concludes by outlining the broader implications of these findings.
\subsection{Model Complexity and Effective Degrees of Freedom}
The performance differences among variable selection methods can be understood through the lens of model complexity, as measured by the effective degrees of freedom (EDF) \citep{efron1986biased}. Defined as $\sigma^{-2}\sum_{i=1}^n\mathrm{cov}(y_i,\hat{y}_i)$, EDF quantifies the ``aggressiveness'' of a fitting procedure.

\begin{figure}[tb]
    \centering
    \includegraphics[width=0.9\linewidth]{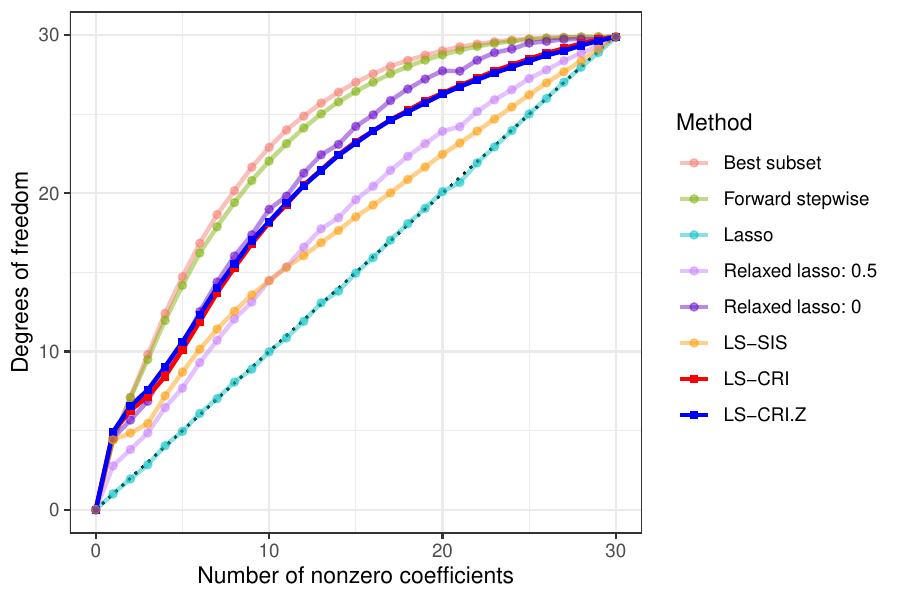}
    \caption{Effective degrees of freedom for the benchmark methods, LS-SIS, and LS-RI variants. Setup mirrors Fig. 4 in \cite{hastie2020best}: beta-type 4, $n=70, p=30, \rho=0.35, s=5, \text{SNR}=0.7$.}
    \label{fig:df}
\end{figure}

As shown in Fig. \ref{fig:df}, the methods occupy distinct regions in the complexity space. Best subset and forward stepwise are the most aggressive, exhibiting the highest EDF for a given model size. The lasso is more conservative, reflecting its bias–variance trade-off. The LS-RI variants are situated between these extremes, offering a balanced level of flexibility. Notably, LS-CRI and LS-CRI.Z follow nearly identical paths, positioning them as a clear midpoint between the aggressive and conservative benchmarks, explaining their balanced performance. In contrast, LS-SIS consistently exhibits lower EDF, reflecting its more conservative behavior, driven by a simpler ranking mechanism.

Fig. \ref{fig:df} also highlights the value of two-parameter methods. The relaxed lasso adapts to varying SNR levels via its tuning parameter $\gamma$, which allows it to interpolate between the lasso and least squares. Similarly, our Ridge-RI variants offer comparable flexibility. By adjusting the ridge penalty $\lambda$, they smoothly control model complexity---from the unregularized LS-RI down to zero degrees of freedom. This adaptive complexity is the key to their superior performance across diverse scenarios, as demonstrated in Section C of the Supplementary Material. 

EDF is inherently data-dependent \citep{tibshirani2015degrees}, and the performance differences observed between LS-CRI and LS-CRI.Z in specific settings can be attributed to such data-dependent shifts in their complexities.

\subsection{Explanatory Fidelity vs. Selection Performance}
A key finding is that the simpler, identity-reallocating RI measures (CRI.Z and CAR) often match or outperform the more elaborate CRI for variable selection and modeling. This outcome is counterintuitive, as CRI provides a more faithful approximation of the theoretical ideal, GD. This distinction suggests a fundamental difference in objective. For explanation, the goal is fidelity. The reallocation step in CRI is critical for equitably distributing importance among correlated predictors. For selection, the goal is discrimination---to robustly separate relevant from irrelevant predictors, where a perfect internal ranking among true predictors is less crucial.

Both CRI and CRI.Z share the minimal transformation of the correlated predictors $X$ into an orthogonal basis $Z$. This transformation is the primary source of robustness to the predictor correlations. CRI.Z uses this signal directly, while CRI adds a reallocation step designed for explanation, which can introduce additional variability into the variable selection task especially in high dimension settings or under strong correlations among variables. The consistent and high performance of CRI.Z across simulation settings suggests that, for variable selection, the minimal transformation is not only sufficient but may be preferable. This opens new research directions for theoretical analysis of CRI.Z and related measures in the domain of variable selection.
\subsection{Limitations}
While our results demonstrate the effectiveness of RI-based variable selection, several limitations warrant discussion. First, our method relies on the linear model with Gaussian errors. Although we demonstrated practical utility on gene expression data where the underlying errors are non-Gaussian, we have not formally evaluated robustness under non-Gaussian noise or heteroscedasticity, the latter being characteristic of longitudinal data. Given that relative importance measures can be easily extended to Generalized Linear Models (GLMs) \cite{tonidandel2010determining}, adapting the RI-based methods to GLMs or Generalized Linear Mixed Models (GLMMs) represents a feasible and promising direction for future research.


Second, while our methods show strong real-world performance compared to other benchmarks, they are primarily developed based on linear models. In many pattern recognition tasks, variables exhibit non-linear interactions that RI measures may not capture without prior feature encoding. Nevertheless, extending RI to non-linear settings is a viable research direction; for instance, \citep{zhao2019revenue} explored the extension of relative importance to Generalized Additive Models (GAMs) with truncated power splines. Integrating such non-linear models with our method would offer a powerful solution for future development to enhance generalizability to complex datasets.
\subsection{Conclusion}
This work bridges the divide between relative importance (RI) analysis and variable selection, establishing that RI measures provide a robust foundation for filter-based selection. By leveraging the minimal transformation, RI-based methods achieve robust signal detection when predictors are highly correlated, a setting that challenges many benchmark approaches. Extensive simulations show that LS-RI methods, particularly LS-CRI.Z, deliver consistently high performance across a wide range of conditions. Furthermore, the regularized Ridge-RI variants provide additional adaptability for model building. The analysis of high-dimensional gene expression datasets further illustrates their practical utility and scalability in real-world applications. These findings position the RI measures not merely as a complementary tool for post-hoc explanation but as a competitive and scalable tool for variable selection. We hope this work motivates broader adoption of RI measures into modern statistical learning.

\section*{CRediT authorship contribution statement}
\textbf{Tien-En Chang}: Conceptualization, Data curation, Investigation, Methodology, Visualization, Validation, Formal analysis, Software, Writing – original draft. \textbf{Argon Chen}: Supervision, Conceptualization, Methodology, Funding acquisition, Writing – review \& editing.

\section*{Declaration of competing interest}
The authors declare that they have no known competing financial interests or personal relationships that could have appeared to
influence the work reported in this paper.

\section*{Data availability}
We have shared the code links in the paper.

\section*{Acknowledgments}
This work was partially supported by the Grant NSTC 106-2221-E-002-153-MY3 from National Science and Technology Council of Taiwan.






\bibliographystyle{elsarticle-num} 
\bibliography{mybibliography}



\end{document}


\maketitle
This supplementary document contains computational complexity analysis and plots from the simulations described in the paper ``Variable Selection using Relative Importance Rankings''. Section \ref{sec:complexity} discusses the computation complexity of RI measures. The plots in Section \ref{sec:part1} precisely follow the simulation part I described in the paper. Section \ref{sec:part2} presents full results of the simulation part II, which compare best subset, forward stepwise, the lasso and the relaxed lasso, LS-SIS and LS-RI variants, and Ridge-SIS and Ridge-RI variants to comparison. Section \ref{sec:non-convex} reviews the non-convex penalized methods and presents their implementation details.

\newpage
\tableofcontents
\newpage
\section{Computational Complexity of RI Measures}\label{sec:complexity}
In this section, we analyze the computational complexity of the discussed RI measures. We assume a high-dimensional setting where $p>n$. The complexity of SVD is $\mathcal{O}(np^2)$ for $n>p$ \cite{li2019tutorial}. While for $p>n$, it can be computed in $\mathcal{O}(n^2p)$ by utilizing $XX^\top$ (the kernel trick).
\subsection{General Dominance (GD)}
GD requires calculating the incremental contribution of each predictor across all possible $2^p-1$ sub-models. Assuming a regular linear model update, the complexity is $\mathcal{O}((np^2+p^3)2^{p})$. This is exponential growth makes GD computationally intractable high-dimensional analysis. 
\subsection{Comprehensive Relative Importance (CRI)}
Instead of the minimal transformation, CRI defines a generalized predictor matrix $Z_G$ that maximally correlated to $X$ as:
\begin{equation}
    Z_G=U_rV_r^\top+U_0CV_0^\top
    \label{eq:generalized Z}
\end{equation}
where $U_0\in\mathbb{R}^{n\times (p-r)}$ and $V_0\in\mathbb{R}^{p\times (p-r)}$ span the left null space and null space of $X$, and $C\in\mathbb{R}^{(p-r)\times(p-r)}$ is any orthogonal matrix. The CRI is then defined as the product of a matrix term (Contribution Reallocation) and a vector term (Contribution of $Z_G$):
$$
D(X) = \underbrace{\left((V_rS_rV_r^\top) \odot (V_rS_rV_r^\top)\right)}_{\text{Contribution Reallocation}} \underbrace{\left((V_rU_r^\top y) \odot (V_rU_r^\top y)\right)}_{\text{Contribution of $Z_G$}}.
$$
While the SVD and the vector term can be computed in $\mathcal{O}(n^2p)$, the matrix term is the computational bottleneck. Constructing the dense $p \times p$ matrix requires $p^2$ inner products of dimension $r$, leading to a complexity of $\mathcal{O}(rp^2)\approx\mathcal{O}(np^2)$. Consequently, CRI scales quadratically with $p$, making it significantly slower than the following approaches.

\subsection{Correlation Adjusted marginal coRrelation (CAR)}
The CAR score is defined as $w = (X^\top X)^{-1/2} X^\top y$ where the inverse square root $(X^\top X)^{-1/2}$ is applied to a shrinkage estimate. Following the optimized implementation in \texttt{care} \cite{zuber2021care}, the computation follows three steps:
\begin{enumerate}
    \item Shrinkage estimation. The optimal shrinkage parameter if estimated using a fast algorithm in \cite{schafer2005shrinkage} that exploited the SVD, requiring $\mathcal{O}(n^2p)$.
    \item Marginal correlation. Computing $X^\top y$ requires $\mathcal{O}(np)$.
    \item Inverse square root. By utilizing the tricks described in \cite{zuber2009gene}, this can be calculated without forming a $p\times p$ matrix, resulting in $\mathcal{O}(n^2p)$ operations.
\end{enumerate}
The total complexity of CAR is $O(n^2p)$. This linear scaling with respect to $p$ makes CAR substantially more efficient than CRI.

\subsection{CRI.Z}
CRI.Z utilizes only the second term of the CRI: $w^2_G=(V_rU_r^\top y)\odot (V_rU_r^\top y)$. This requires the SVD of $X$ ($\mathcal{O}(n^2p)$) followed by matrix-vector multiplication ($\mathcal{O}(np)$). Like CAR, CRI.Z belongs to the $O(n^2p)$ complexity class. In practice, CRI.Z is marginally faster than CAR as it avoids the computational overhead associated with shrinkage parameter estimation. Both measures are significantly faster than CRI due to bypassing the construction of $p\times p$ matrices.
\newpage
\section{Simulation Part I}\label{sec:part1}
\subsection{Low setting: $n=100$, $p=10$, $s=5$} 
\subsubsection{Minimal Model Size $(S)$} 
\begin{figure}[!h]
\centering
\includegraphics[width=1\linewidth]{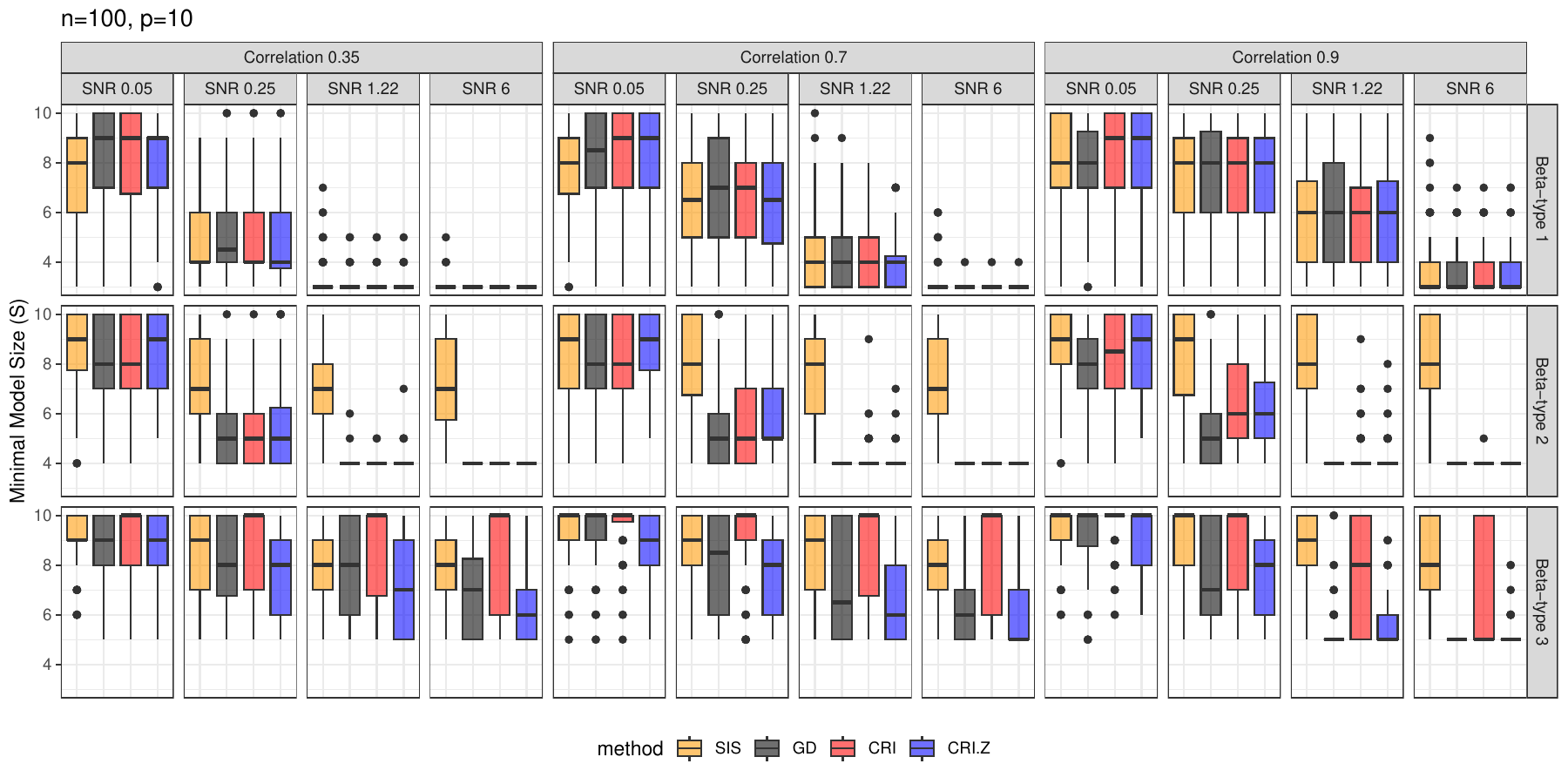}
\end{figure}

\subsubsection{Selected Proportion $Pr(k)$} 
\begin{figure}[!h]
\centering
\includegraphics[width=1\linewidth]{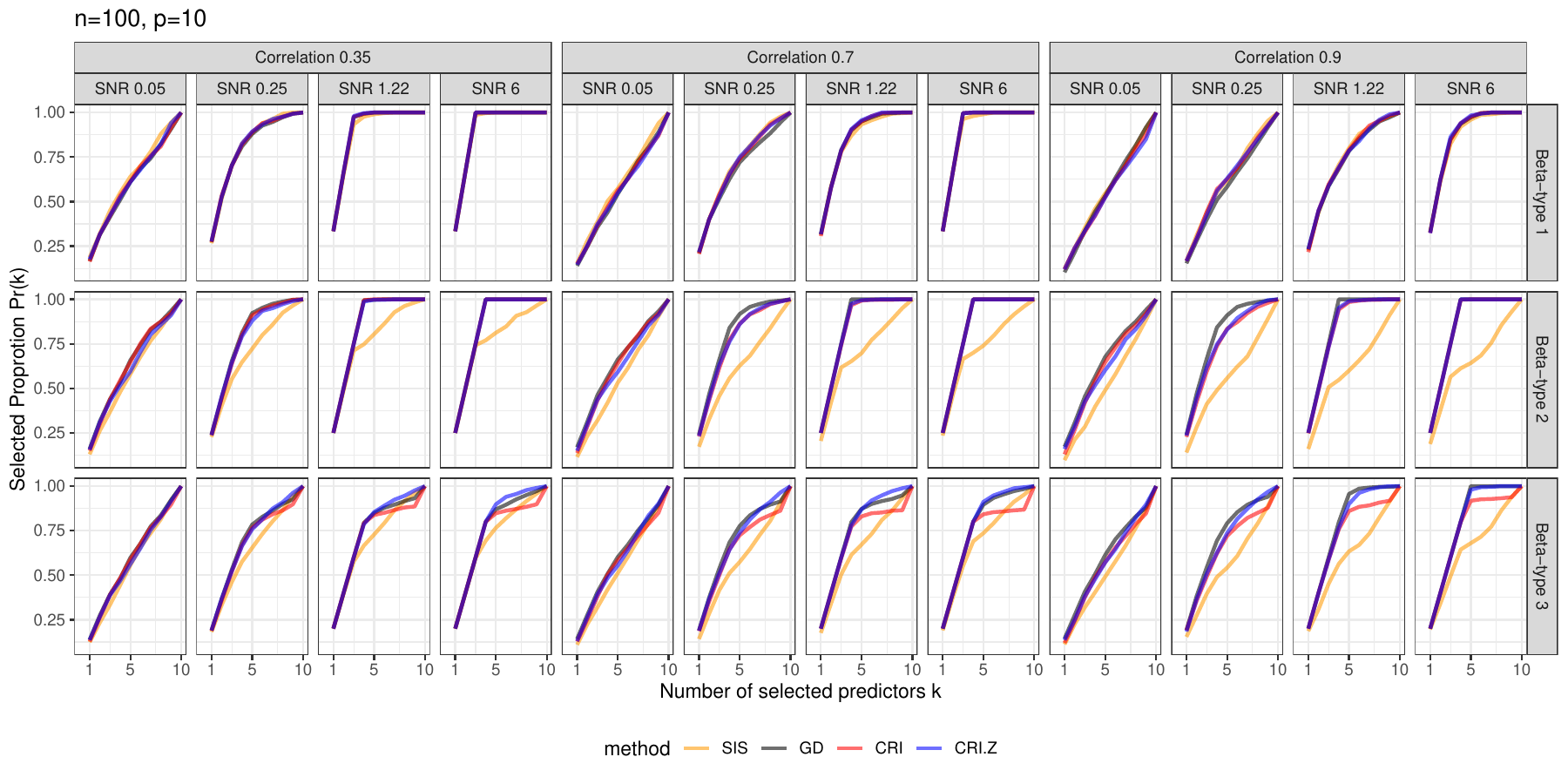}
\end{figure}
\newpage

\subsection{Medium setting: $n=500$, $p=100$, $s=5$} 
\subsubsection{Minimal Model Size $(S)$} 
\begin{figure}[!h]
\centering
\includegraphics[width=1\linewidth]{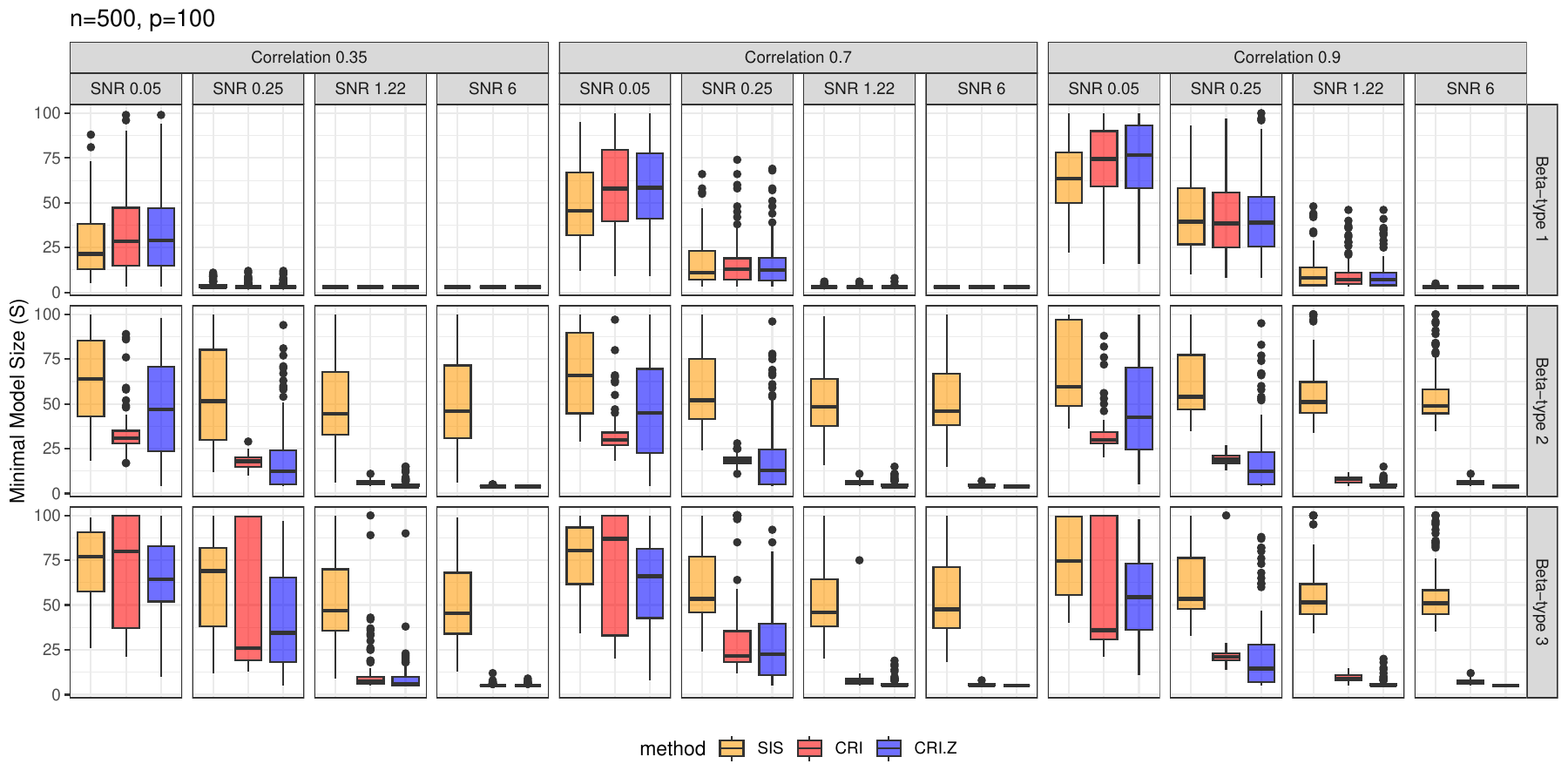}
\end{figure}

\subsubsection{Selected Proportion $Pr(k)$} 
\begin{figure}[!h]
\centering
\includegraphics[width=1\linewidth]{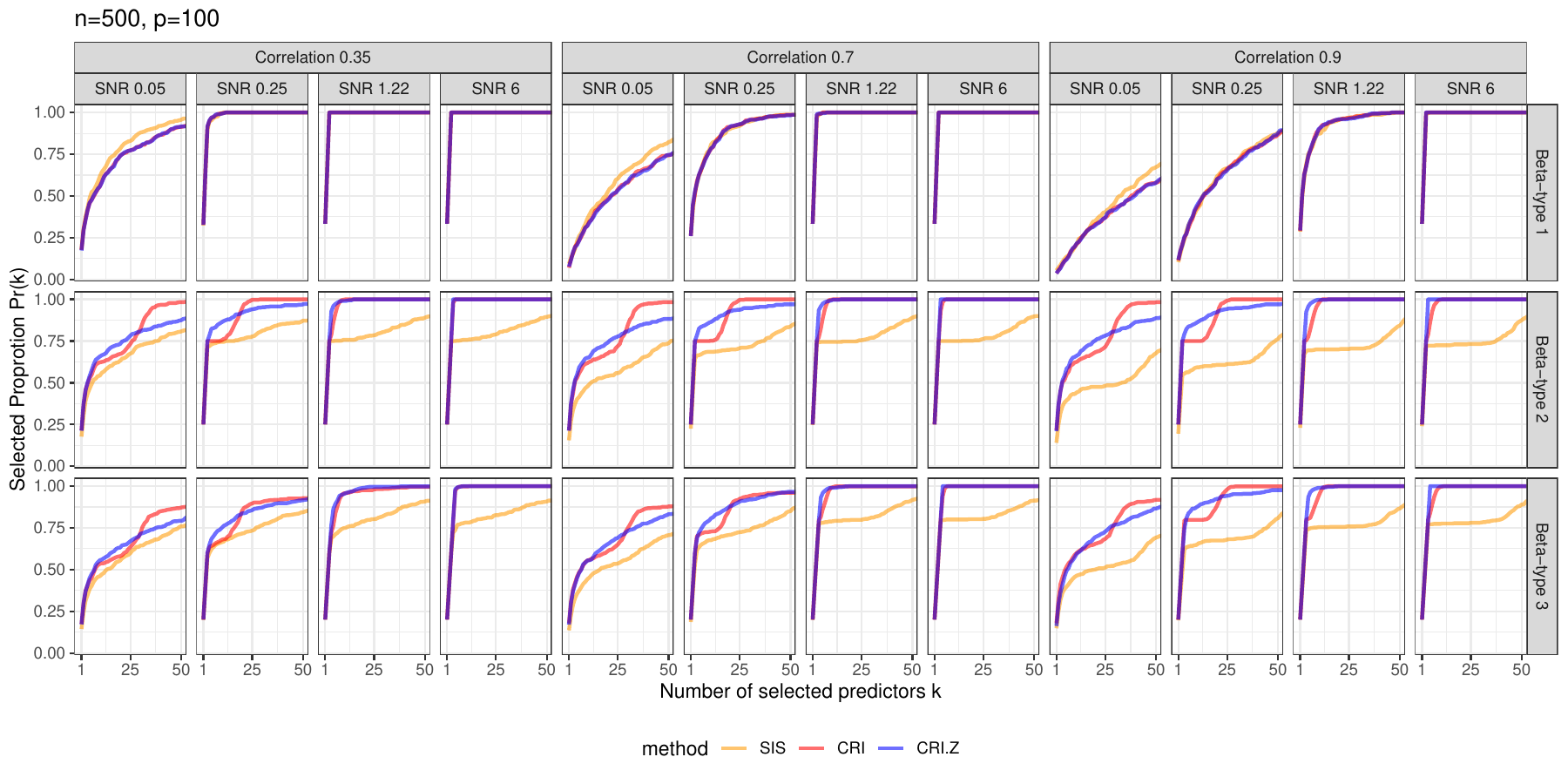}
\end{figure}
\newpage

\subsection{High-50 setting: $n=50$, $p=1000$, $s=5$} 
\subsubsection{Minimal Model Size $(S)$} 
\begin{figure}[!h]
\centering
\includegraphics[width=1\linewidth]{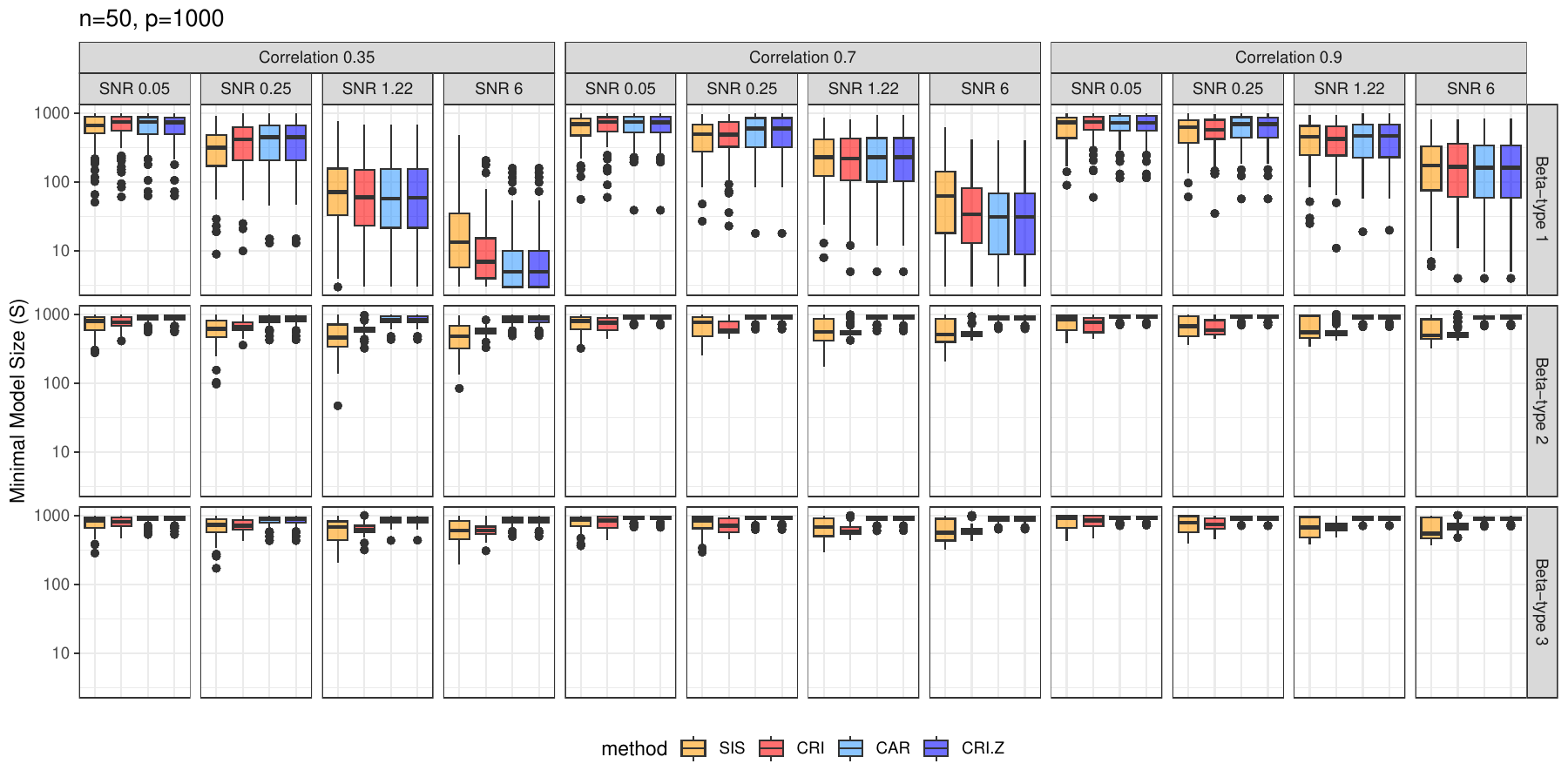}
\end{figure}

\subsubsection{Selected Proportion $Pr(k)$} 
\begin{figure}[!h]
\centering
\includegraphics[width=1\linewidth]{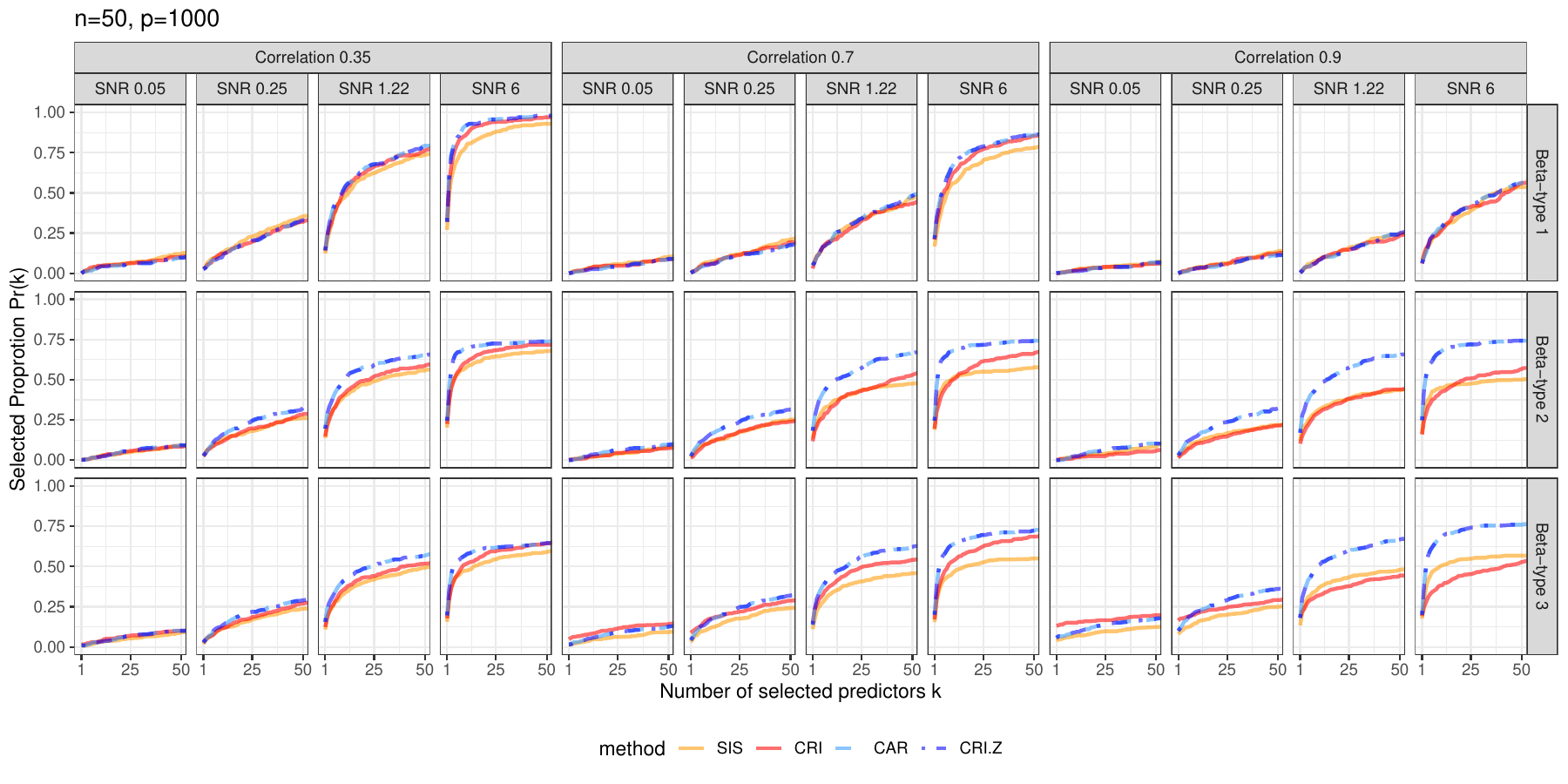}
\end{figure}
\newpage

\subsection{High-100 setting: $n=100$, $p=1000$, $s=5$} 
\subsubsection{Minimal Model Size $(S)$} 
\begin{figure}[!h]
\centering
\includegraphics[width=1\linewidth]{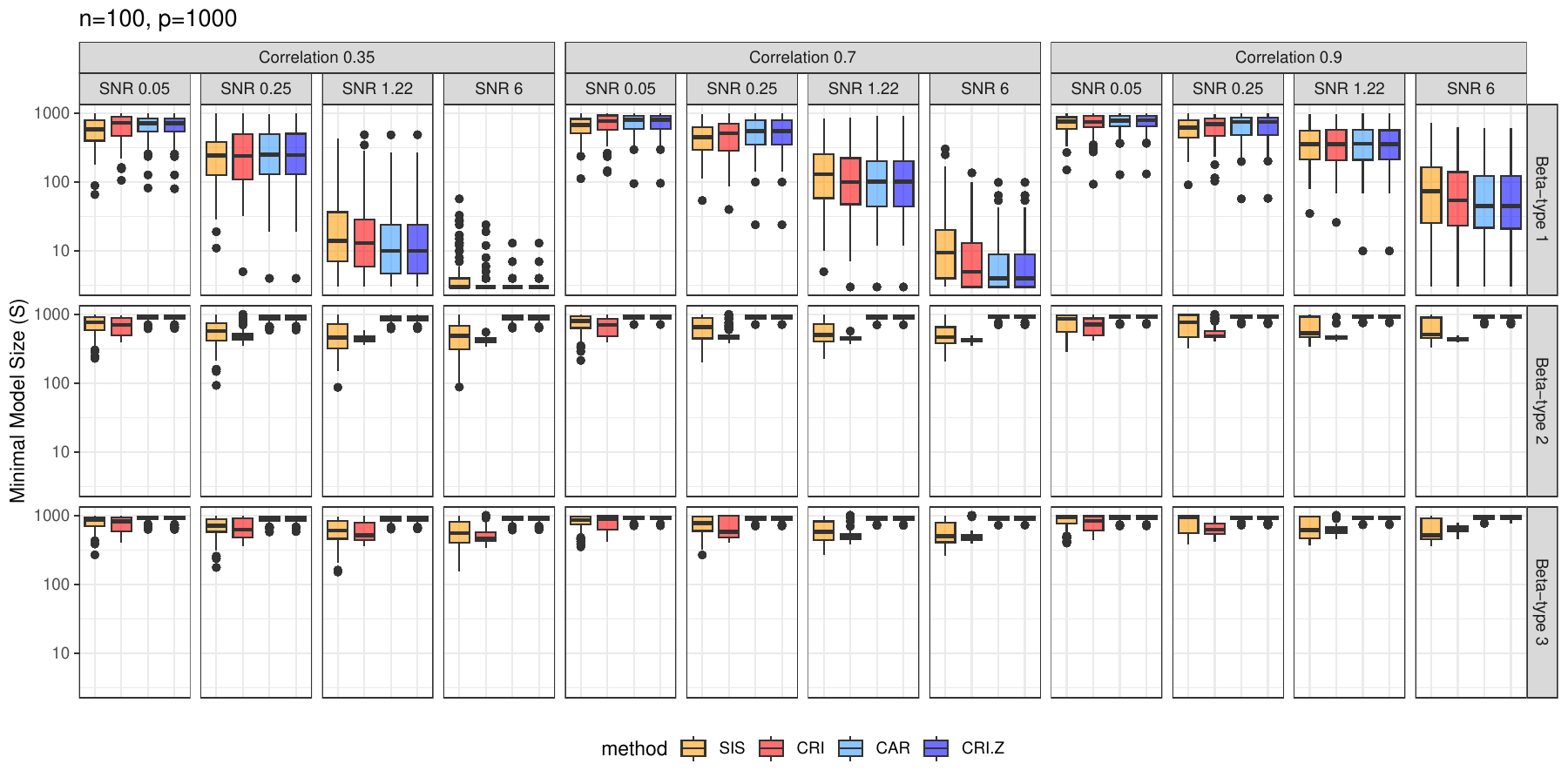}
\end{figure}

\subsubsection{Selected Proportion $Pr(k)$} 
\begin{figure}[!h]
\centering
\includegraphics[width=1\linewidth]{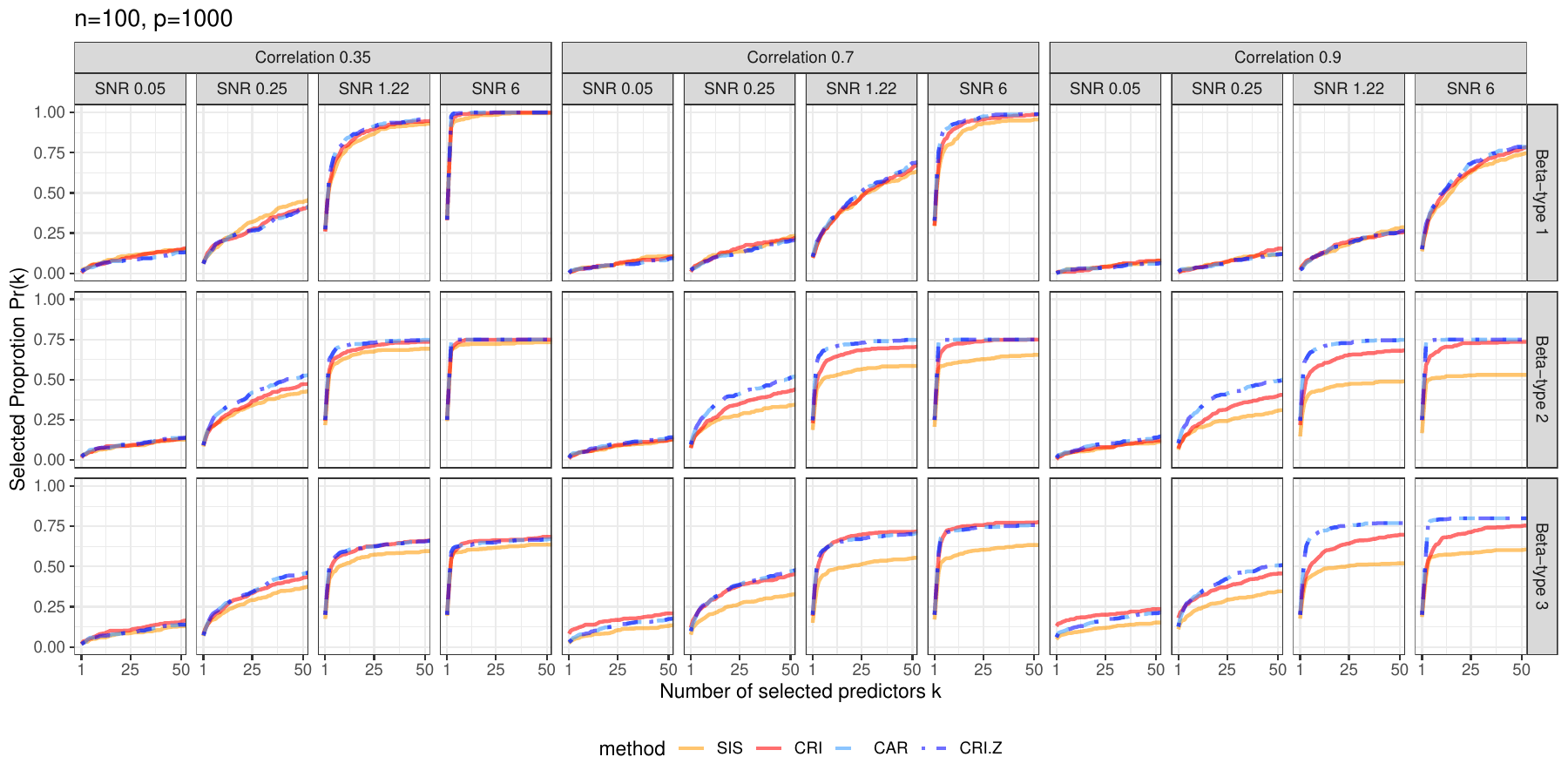}
\end{figure}
\newpage
\section{Simulation Part II}\label{sec:part2}
The general RI-based selection algorithm is outlined in Algorithm~\ref{alg:RI-based unified}. 
\begin{algorithm}[tb]
    \begin{algorithmic}[1]
    \Require Predictor matrix $X$, response vector $y$, RI measure $f(\cdot)$, max model size $K$. (Optional: tuning parameters $\{\lambda_1,...,\lambda_M\}$ for Ridge-RI.)
    \Ensure A sequence of fitted models $\hat{\beta}_{A_1}, ..., \hat{\beta}_{A_K}$.
    \State Compute the relative importance measures for all predictors $d=f(X,y)$.
    \State Initialize the active set $A_0=\{\}$.
    \For {$k\in\{1,...,K\}$}
        \State $i=\argmax_{j\notin A_{k-1}} d_j$
        \State $A_k \gets A_{k-1}\cup\{i\}$
        \State \textit{// Model 1: LS-RI variants}
        \State $\hat{\beta}_{A_k}^{\mathrm{LS}}\gets (X^\top_{A_k}X_{A_k})^{-1}X^\top_{A_k}y$
        \State \textit{// Model 2: Ridge-RI variants}
        \For {$\lambda\in\{\lambda_1,...,\lambda_M\}$}
            \State $\hat{\beta}_{A_k}^{\mathrm{ridge}}(\lambda)\gets (X^\top_{A_k}X_{A_k}+\lambda I)^{-1}X^\top_{A_k}y$
        \EndFor
    \EndFor
    \State \Return The sequence(s) of all fitted models.
    \caption{The RI-based Variable Selection Methods}
    \label{alg:RI-based unified}
    \end{algorithmic}
\end{algorithm}

We implemented the ridge regression component of the Ridge-SIS and Ridge-RI variants using the \texttt{glmnet} package. The tuning process involved a joint grid search over two hyperparameters: subset size $k$ and ridge penalty $\lambda$.

\begin{itemize}
    \item \textit{Grid Specification.} In the low-dimension setting, we tuned over $k \in \{0, \dots, 10\}$ and used a sequence of 10 $\lambda$ values. In all other settings (medium, high-50, high-100), we tuned over $k \in \{0, \dots, 50\}$ and used a sequence of 20 $\lambda$ values. This design results in a total hyperparameter search space (e.g., $51 \times 20 \approx 1000$) that is directly comparable to the grid size used for the relaxed lasso benchmark ($100 \times 10 = 1000$).

    \item \textit{$\lambda$ Sequence Generation.} For each candidate active set determined by $k$, the $\lambda$ sequence is generated using the standard strategy of \texttt{glmnet}. Values are spaced linearly on a log scale from \texttt{lambda.max} down to \texttt{lambda.min}. Since the theoretical \texttt{lambda.max} for pure ridge regression ($\alpha=0$) is infinite, \texttt{glmnet} approximates this value by computing the smallest penalty that shrinks coefficients to zero under a $\alpha$ sufficiently close to zero. The lower bound \texttt{lambda.min} is then determined by the ratio \texttt{lambda.min.rate} (\texttt{lambda.min}/\texttt{lambda.max}). Following the defaults, we set this ratio to $0.01$ if $p>n$ and $10^{-4}$ otherwise.
\end{itemize}
It is important to note that the grids for regularized methods (lasso, relaxed lasso, and Ridge-RI) strictly evaluate regularized solutions (i.e., $\lambda > 0$). 

\newpage
\subsection{Low setting: $n=100$, $p=10$, $s=5$} 
\subsubsection{F1-score} 
\begin{figure}[!h]
\centering
\includegraphics[width=1\linewidth]{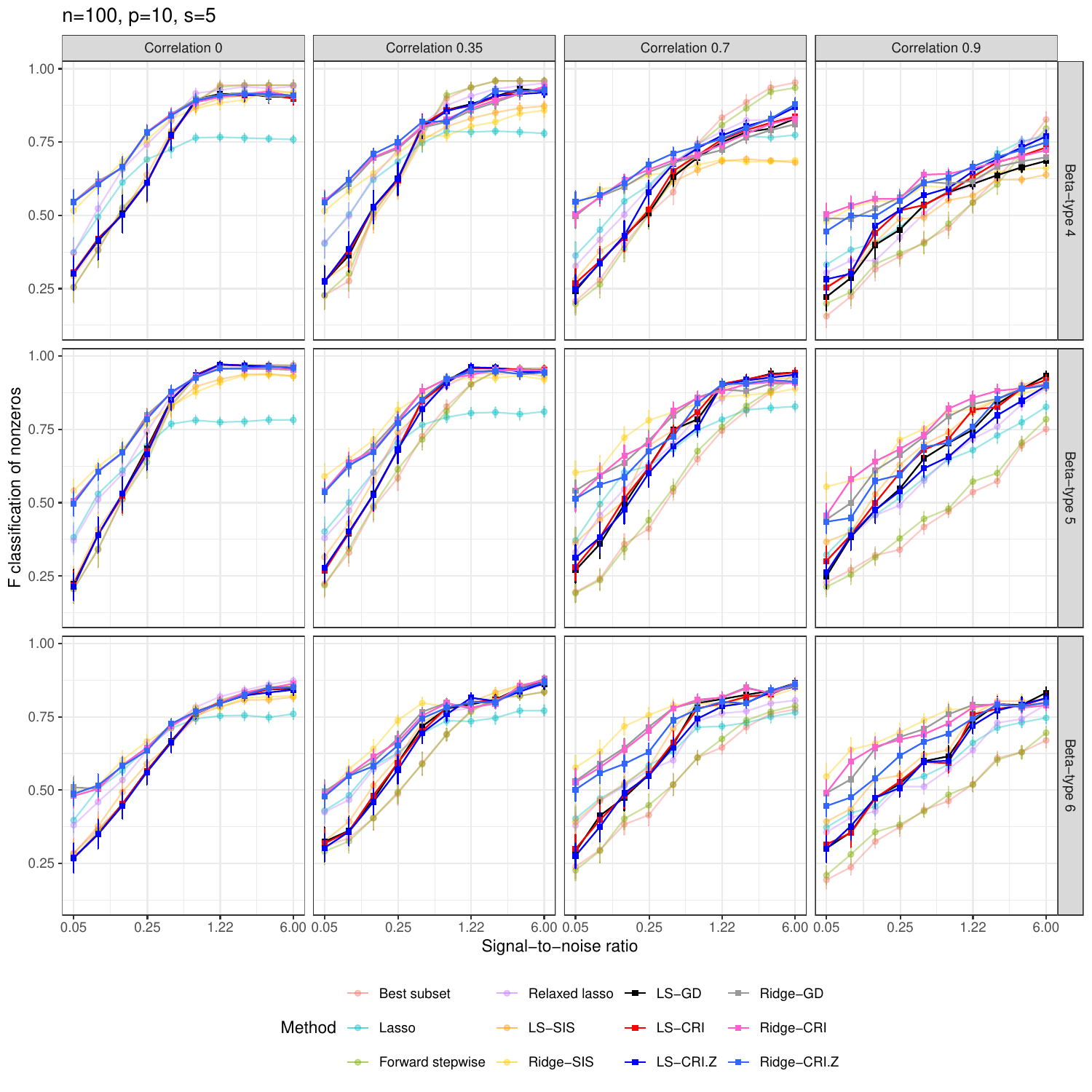}
\end{figure}
\newpage

\subsubsection{Relative test error (to Bayes)}
\begin{figure}[!h]
\centering
\includegraphics[width=1\linewidth]{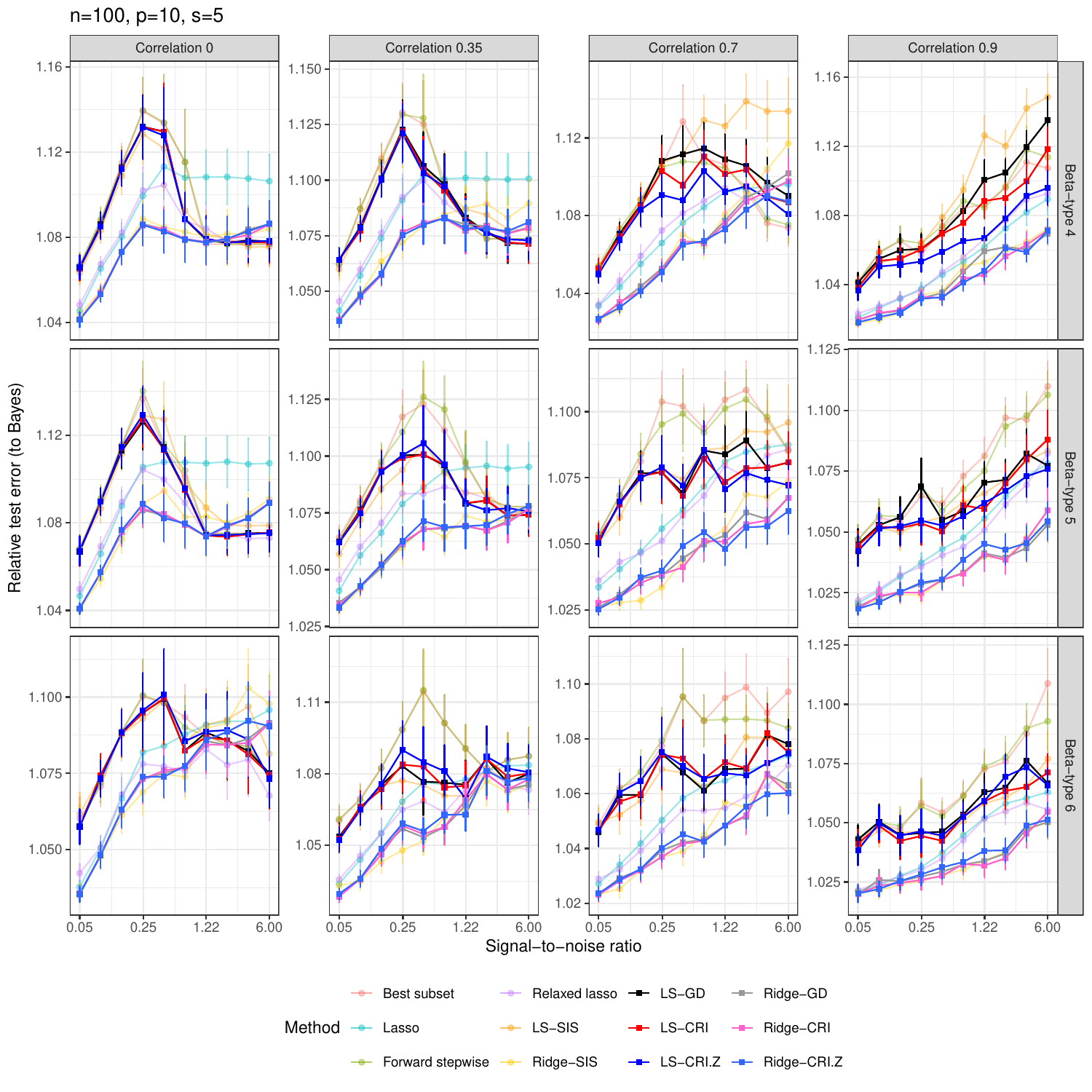}
\end{figure}
\newpage

\subsection{Medium setting: $n=500$, $p=100$, $s=5$} 
\subsubsection{F1-score} 
\begin{figure}[!h]
\centering
\includegraphics[width=1\linewidth]{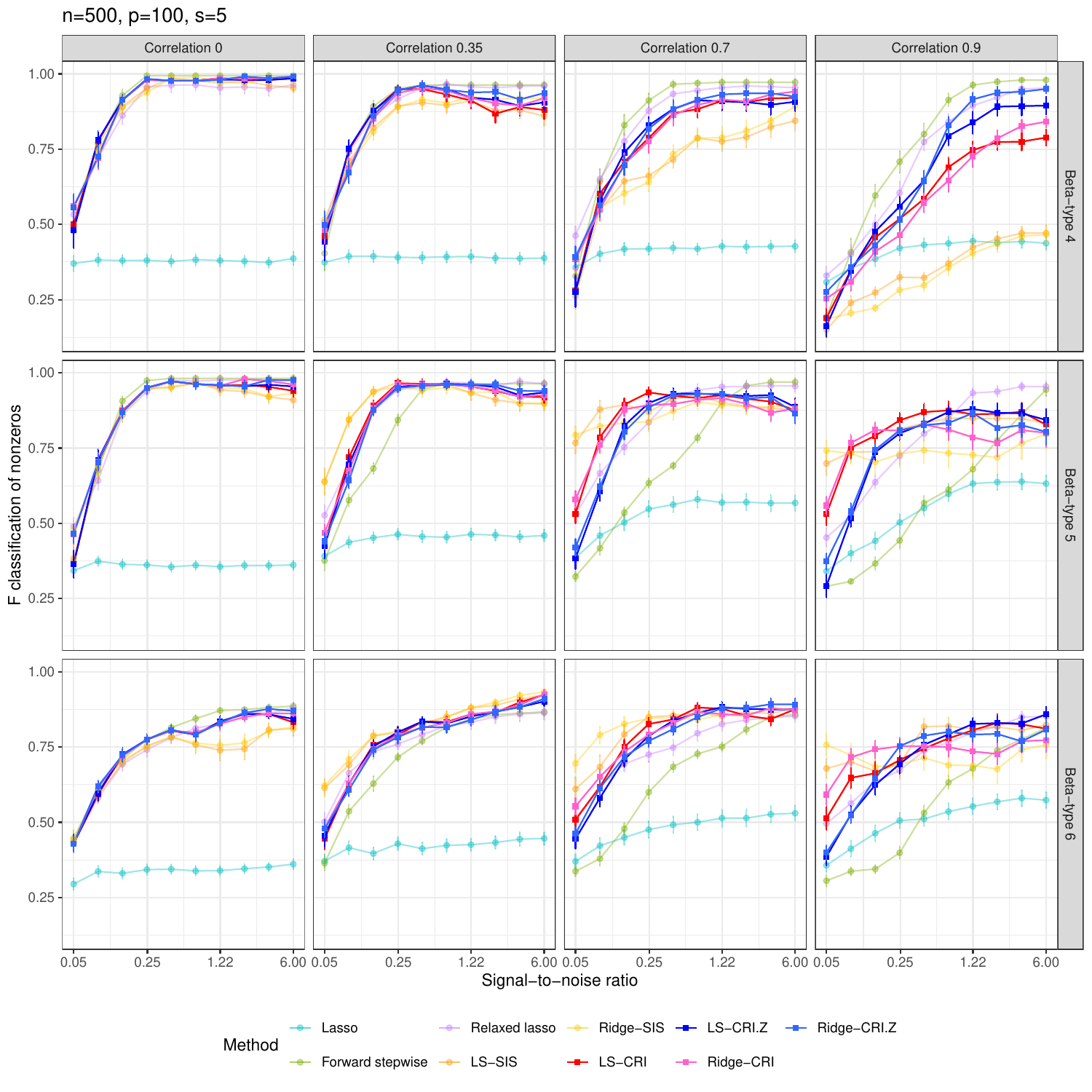}
\end{figure}
\newpage

\subsubsection{Relative test error (to Bayes)}
\begin{figure}[!h]
\centering
\includegraphics[width=1\linewidth]{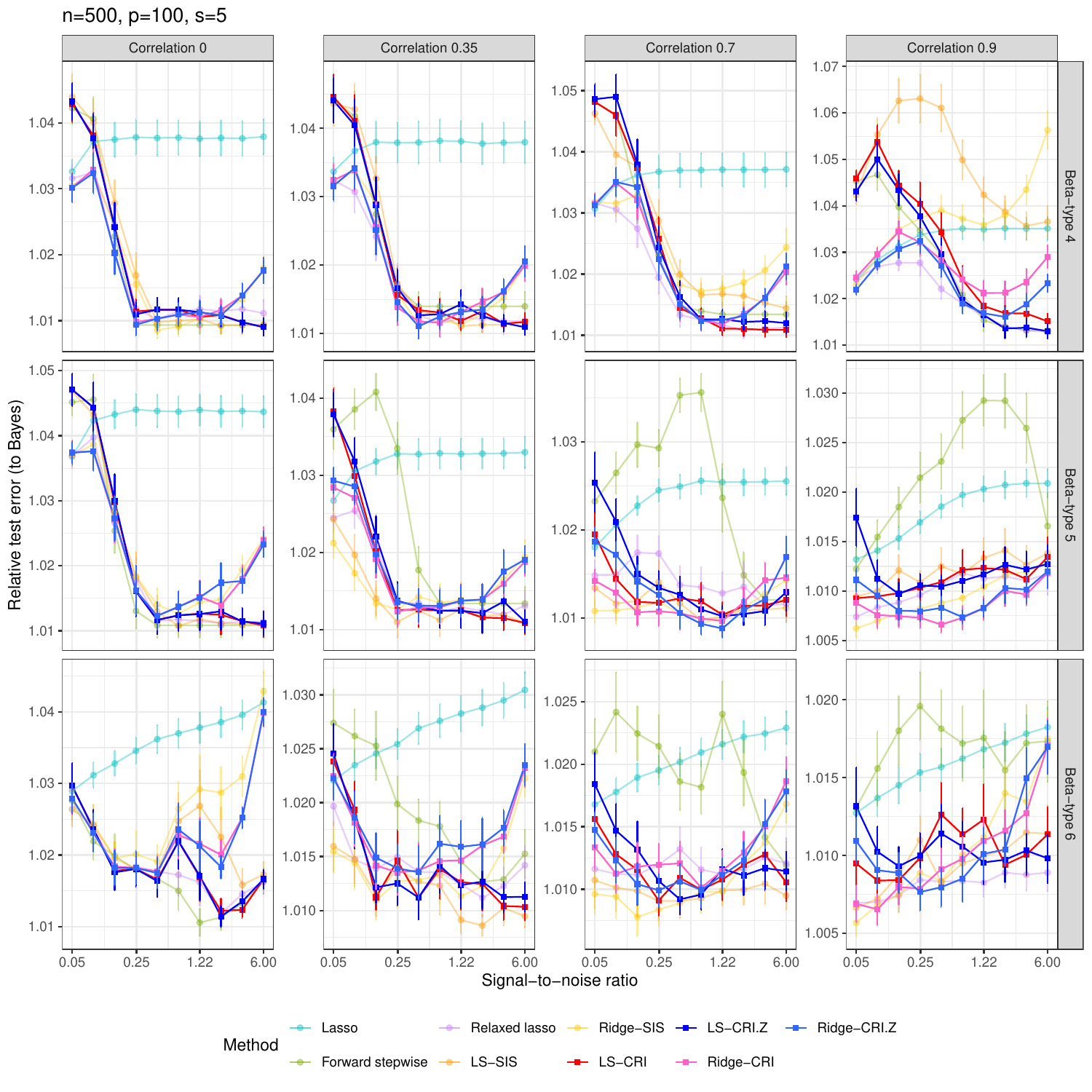}
\end{figure}
\newpage

\subsection{High-50 setting: $n=50$, $p=1000$, $s=5$} 
\subsubsection{F1-score} 
\begin{figure}[!h]
\centering
\includegraphics[width=1\linewidth]{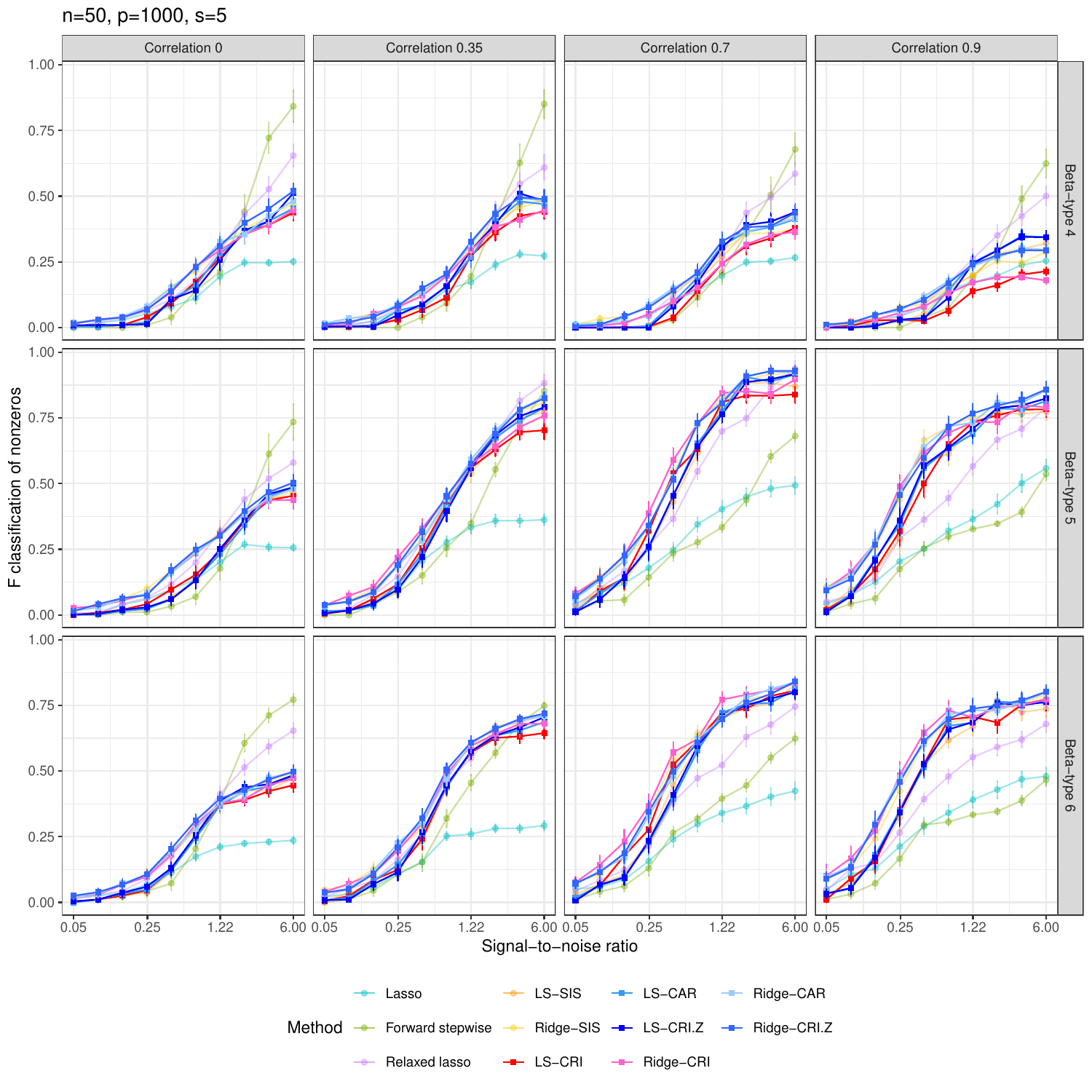}
\end{figure}
\newpage

\subsubsection{Relative test error (to Bayes)}
\begin{figure}[!h]
\centering
\includegraphics[width=1\linewidth]{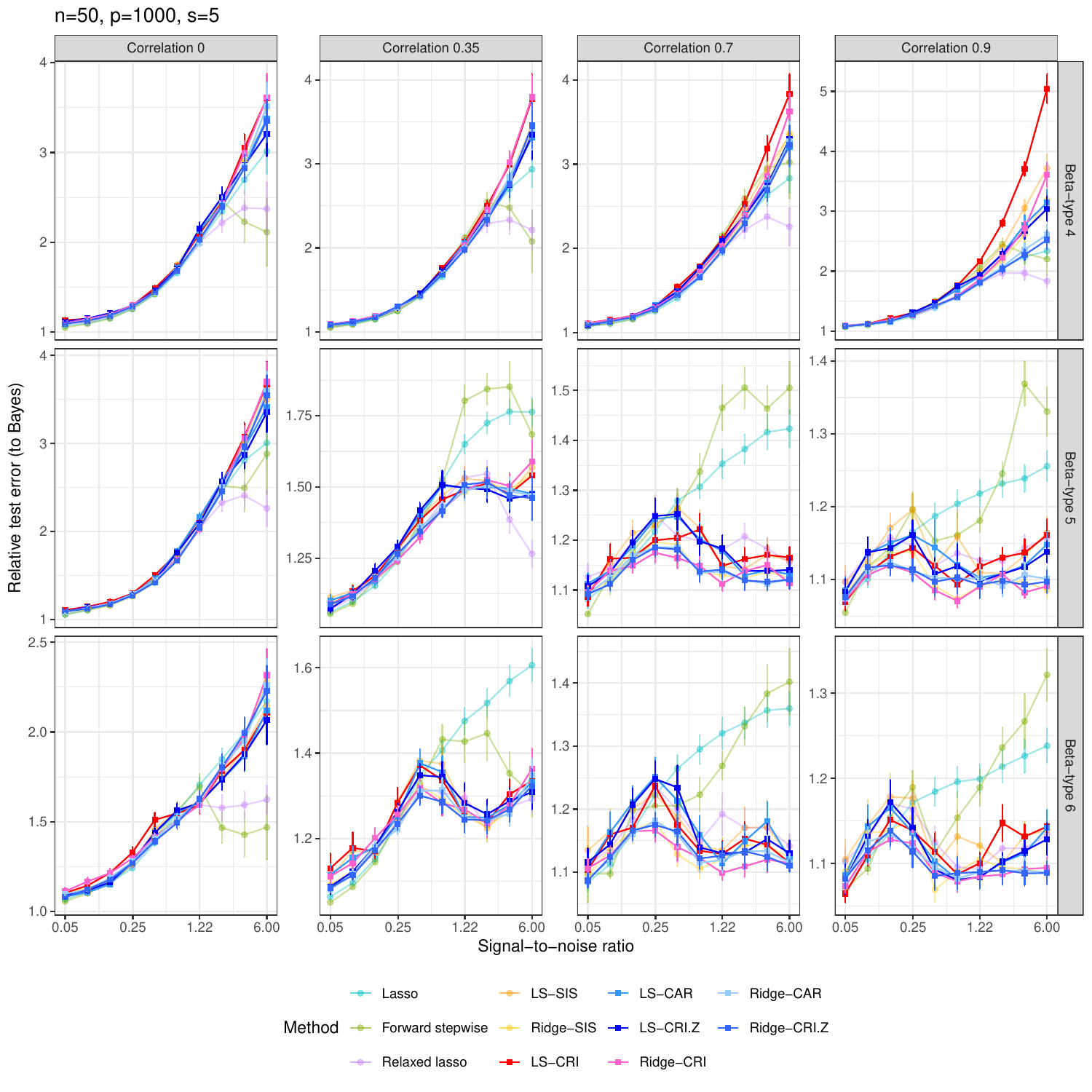}
\end{figure}
\newpage

\subsection{High-100 setting: $n=100$, $p=1000$, $s=5$} 
\subsubsection{F1-score} 
\begin{figure}[!h]
\centering
\includegraphics[width=1\linewidth]{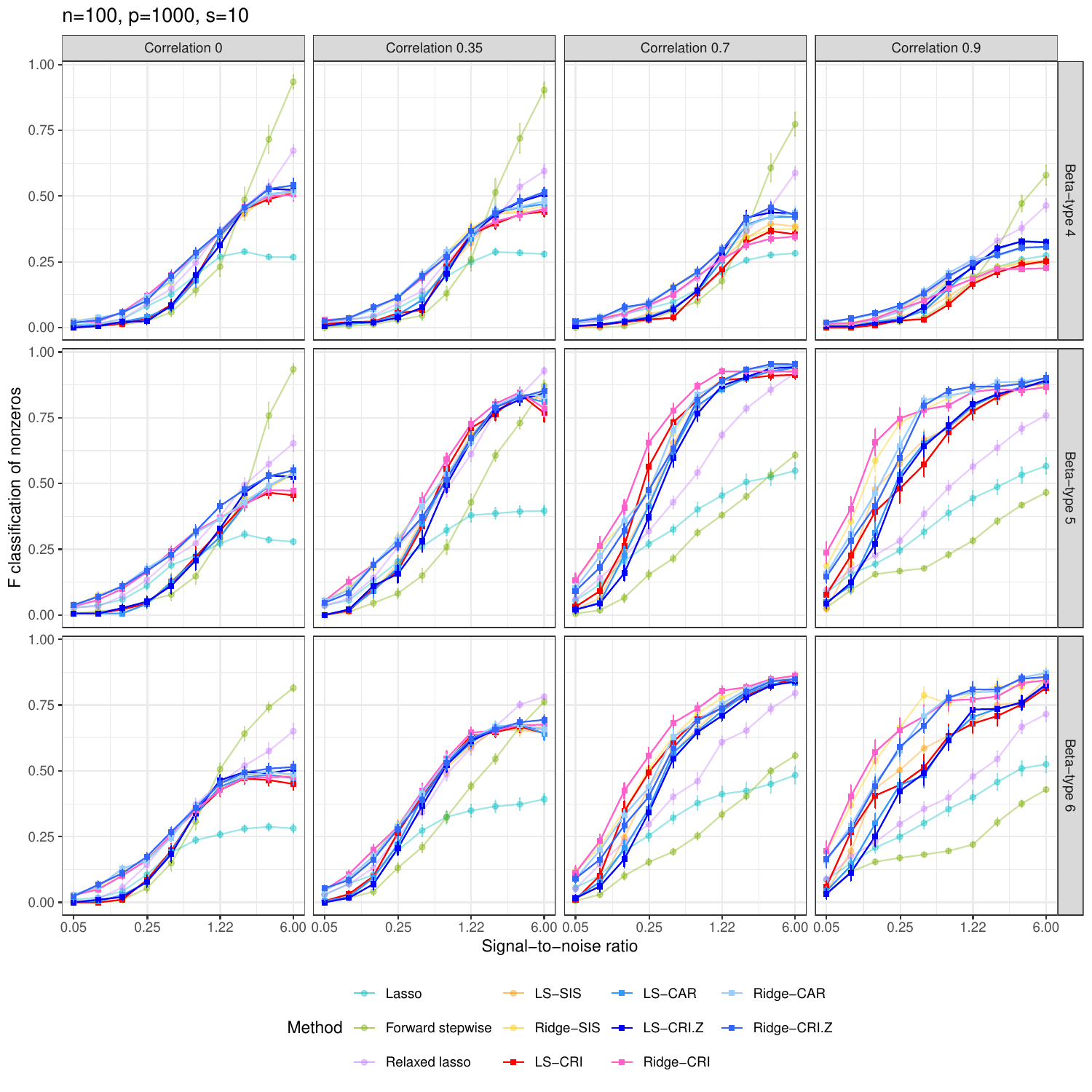}
\end{figure}
\newpage

\subsubsection{Relative test error (to Bayes)}
\begin{figure}[!h]
\centering
\includegraphics[width=1\linewidth]{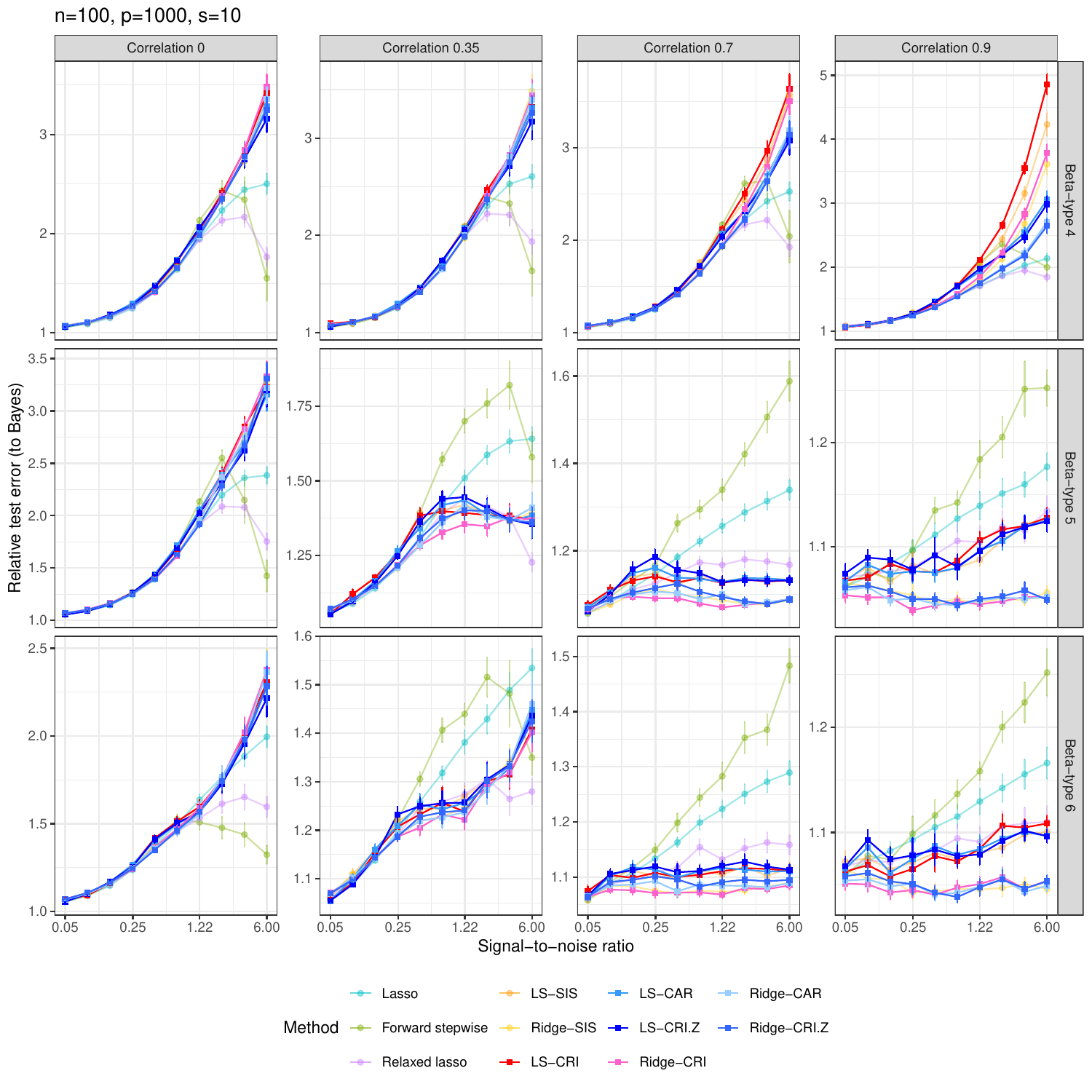}
\end{figure}
\newpage

\subsection{Runtime}
Table~\ref{tab:runtime} summarizes the average computational runtimes in simulation part II. 
\begin{table}[h]
    \centering
    \renewcommand{\thetable}{S1}
    \caption{Average computational runtime (in milliseconds).}
    \label{tab:runtime}
    \begin{tabular}{lrrrr} 
    \hline
    \multirow{2}{*}{Method} & \multicolumn{2}{c}{$n>p$} & \multicolumn{2}{c}{$p>n$}\\     
    \cmidrule(r){2-3} \cmidrule(l){4-5}
     & low & medium & high-50 & high-100 \\ \hline
    \textit{Benchmarks} & & & & \\
    Best subset & 204.0 & -- & -- & -- \\
    Lasso & 0.9 & 3.6 & 3.6 & 6.1 \\
    Forward stepwise & 2.0 & 187.9 & 78.1 & 141.5 \\
    Relaxed lasso & 0.9 & 4.0 & 2.9 & 5.2 \\
    LS-SIS & 0.8 & 12.7 & 3.7 & 4.9 \\ \hline
    \textit{RI-Based Methods} & & & & \\
    LS-GD & 26.9 & -- & -- & -- \\
    LS-CRI & 1.1 & 24.2 & 23.5 & 53.3 \\
    LS-CAR & 1.1 & 24.2 & 12.5 & 37.4 \\
    LS-CRI.Z & 1.0 & 23.9 & 8.3 & 24.2 \\
    Ridge-CRI.Z & 5.7 & 49.9 & 71.0 & 83.8 \\ \hline
    \multicolumn{5}{l}{\footnotesize -- : Computationally intractable.} \\
    \end{tabular}%
\end{table}

\section{Details of Non-convex Penalized Methods}\label{sec:non-convex}
To address the estimation bias of the lasso, besides the relaxed lasso, we also include two prominent non-convex penalized regression methods, SCAD \cite{fan2001variable} and MCP \cite{zhang2010nearly}, as benchmarks in our real-world dataset analysis. Both methods solve an optimization problem of the form:
\begin{equation}
    \label{eq:prob_penalized_supp}
    \min_{\beta\in\mathbb{R}^p}  \frac{1}{2}\norm{y-X\beta}^2_2 + \sum_{j=1}^p g(\beta_j; \lambda, \gamma) ,
\end{equation}
where $g(\cdot)$ is the penalty function. The SCAD penalty is defined as:
\begin{equation}
g^{\mathrm{SCAD}}(x; \lambda, \gamma) = \begin{cases} 
    \lambda|x| & \text{if } |x| \leq \lambda, \\
    \frac{2\gamma\lambda|x| - x^2 - \lambda^2}{2(\gamma - 1)} & \text{if } \lambda < |x| \leq \gamma\lambda, \\
    \frac{\lambda^2(\gamma + 1)}{2} & \text{otherwise,}
\end{cases}
\end{equation}
for $\gamma > 2$, while the MCP penalty is defined as:
\begin{equation}
g^{\mathrm{MCP}}(x; \lambda, \gamma) = \begin{cases} 
    \lambda|x| - \frac{x^2}{2\gamma} & \text{if } |x| \leq \gamma\lambda, \\
    \frac{1}{2}\gamma\lambda^2 & \text{otherwise,}
\end{cases}
\end{equation}
for $\gamma > 1$.

We implemented SCAD and MCP using the \texttt{ncvreg} \texttt{R} package \cite{breheny2011coordinate}. To ensure a fair comparison, we conducted a grid search to jointly tune the regularization parameter $\lambda$ and the convexity parameter $\gamma$. 

\begin{itemize}
    \item \textit{$\lambda$ Sequence}: We used a sequence consistent with the range used for the lasso and relaxed lasso.
    \item \textit{$\gamma$ Sequence}: Following the default strategy in \texttt{ncvreg}, we evaluated $\texttt{ngamma} = 9$ values spaced linearly on a log scale from $\texttt{max.gamma} = 150$ down to $\texttt{min.gamma}$. 
    \item \textit{Parameter Constraints}: For SCAD, we set $\texttt{min.gamma} = 2.001$, and for MCP, $\texttt{min.gamma} = 1.001$, ensuring the penalties remain well-defined and satisfy their respective theoretical constraints.
\end{itemize}
The final model was selected based on the combination of $(\lambda, \gamma)$ that achieved the lowest error on the validation set.

\newpage
\bibliographystyle{abbrv} 
\bibliography{references}